\documentclass{article}

\PassOptionsToPackage{numbers}{natbib}

\usepackage[preprint]{neurips_2021}

\usepackage[utf8]{inputenc} 
\usepackage[T1]{fontenc}    
\usepackage{hyperref}       
\usepackage{url}            
\usepackage{booktabs}       
\usepackage{amsfonts}       
\usepackage{nicefrac}       
\usepackage{microtype}      
\usepackage{xcolor}         

\usepackage{graphicx}
\usepackage{amsmath}
\usepackage{xspace}
\usepackage{subcaption}
\newcommand{\WIP}[1]{}
\newcommand{\note}[1]{}
\newcommand{\AR}[1]{}

\newcommand{\tea}{\text{t}}
\newcommand{\stu}{\text{s}}
\newcommand{\proxy}{\text{proxy}}
\newcommand{\down}{\text{d}}
\newcommand{\ie}{\emph{i.e.\@}\xspace}
\newcommand{\eg}{\emph{e.g.\@}\xspace}
\newcommand{\etal}{\emph{et~al.\@}\xspace}

\newcommand{\etc}{\emph{etc.\@}\xspace}
\newcommand{\vs}{\emph{vs.\@}\xspace}
\newcommand{\Misc}{Misc.\@\xspace}

\makeatletter
\newcommand*{\storecounter}[2]{%
  \edef\@currentlabel{\the\value{#1}}
  \label{#2}
}
\makeatother

\title{Representation Consolidation \\ for Training Expert Students} 

\author{\textbf{Zhizhong Li \quad Avinash Ravichandran \quad Charless Fowlkes} \\ \textbf{Marzia Polito \quad Rahul Bhotika \quad Stefano Soatto} \\
Amazon AWS Rekognition \\
  \texttt{\{lzhizhon,ravinash,fowlkec,mpolito,bhotikar,soattos\}@amazon.com} \\}
  
\begin{document}

\maketitle
\begin{abstract}
    Traditionally, distillation has been used to train a student model to emulate the input/output functionality of a teacher. A more useful goal than emulation, yet under-explored, is for the student to learn feature representations that transfer well to future tasks. However, we observe that standard distillation of task-specific teachers actually {\em reduces} the transferability of student representations to downstream tasks. 
    We show that a  multi-head, multi-task distillation method using an unlabeled proxy dataset and a generalist teacher is sufficient to consolidate representations from task-specific teacher(s) and improve downstream performance, outperforming the teacher(s) and the strong baseline of ImageNet pretrained features. Our method can also combine the representational knowledge of multiple teachers trained on one or multiple domains into a single model, whose representation is improved on all teachers' domain(s).
\end{abstract}

\section{Introduction}
\label{sec:intro}

A promising approach to scale up transfer learning to diverse downstream vision tasks is to maintain a library of pre-trained experts. When presented with a novel downstream task, one can select an appropriate expert and quickly fine-tune  the representation with small amounts of task-specific data. This strategy has many practical benefits: fine-tuning a task-relevant pre-trained representation is fast, there is no need to store or revisit expert pre-training data, and in principle the system can be ``upgraded'' at any time by simply adding additional experts to the library. How is such a library of expert representations populated? Previous work (e.g., \cite{puigcerver2020scalable,achille2019task2vec,deshpande2021linearized}\WIP{,others}) has assumed a static collection of experts created by training on domain-specific datasets or semantically related subsets of classes selected from large-scale general-purpose datasets. Instead, we would like to automatically maintain and enrich the library of experts based on the accumulated experience of tuning models for downstream tasks, so that the overall system performance continually increases over time (life-long meta-learning). 

One approach to growing the library would be to add every fine-tuned downstream task-specific model back into the library as a candidate expert for transfer on future tasks. Such a na\"ive approach clearly does not scale and, at a minimum, requires developing techniques for selecting experts that is sub-linear in the size of the library~\cite{puigcerver2020scalable,achille2019task2vec,deshpande2021linearized}\WIP{,others}.  More importantly, as our experiments show, task-specific models fine-tuned on small amounts of data do not provide transferable representations.  Task-specific models tend to overspecialize, suffering ``catastrophic forgetting'' of under-utilized feature representations and thus under-perform on new tasks compared to generic pre-trained models. To address this, we introduce {\em representation consolidation} in which the goal is to consolidate knowledge from multiple task-specific teacher models into a single expert student representation which transfers to downstream tasks {\em better} than any of the individual teacher representations.

\begin{figure}
    \centering
    \includegraphics[width=0.7\columnwidth]{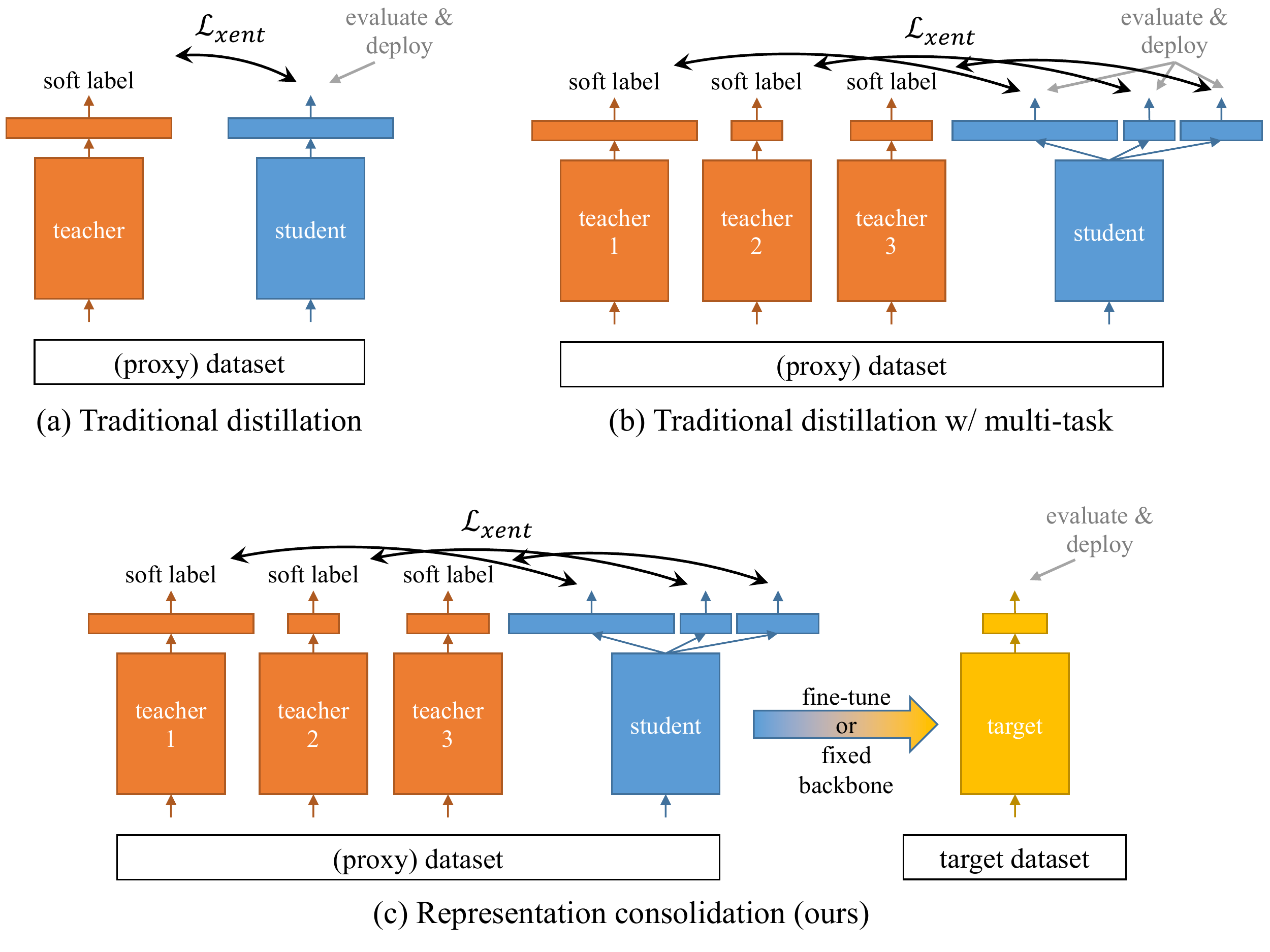}
    \caption{Unlike knowledge distillation, which seeks to copy the end-to-end functionality a teacher, we learn a consolidated representation from multiple teachers that transfers to downstream tasks. Given $N$ teacher models and a large unlabeled proxy dataset, we train a single student model using multi-task distillation with a separate classifier head for each of the teacher tasks. To limit student forgetting and representation collapse, we always include an additional generalist teacher (ImageNet). We show that the resulting consolidated representation transfers better to downstream tasks than any of the individual teachers (including the generalist).}
    \label{fig:method_illust}
\end{figure}

To carry out representation consolidation, we utilize multi-teacher multi-task model distillation (see Fig.\ref{fig:method_illust}). Here, a single student is trained to emulate multiple teachers, each of which operates on a different set of class labels. Previous work on multi-teacher knowledge distillation has focused on evaluating how well the student model performs the teacher's task. Instead, we evaluate how well the student representation generalizes to new downstream tasks (whether related or unrelated to the teachers' tasks). In this setting we demonstrate several surprising results:
\begin{itemize}
    \item While task-specific model representations transfer poorly, consolidating a task-specific teacher with a generalist teacher (ImageNet) is sufficient to rescue the student. The resulting representation transfers well, with improved downstream performance on teacher-relevant tasks while matching the  performance of a strong generalist representation on unrelated tasks.
    \item Consolidating multiple related task-specific teacher models can yield a student representation that exceeds the performance of any one teacher on  downstream tasks.
    \item Unlike knowledge distillation, which requires access to the teacher training data (or using data-free distillation \cite{Lopes2017DataFreeKD,Luo2020LargeScaleGD} to carefully craft synthetic data) to achieve good performance, we avoid using these data and show effective representation consolidation can be carried out using a sufficiently diverse generic proxy dataset and is robust to the choice of the proxy.
\end{itemize}   

%
%


\begin{table}[b]
    \centering
    {\renewcommand{\arraystretch}{1.1}
    \resizebox{0.49\columnwidth}{!}{
    \begin{tabular}{cl} \toprule
        Symbol & Definition \\ \midrule
        $N$ & The number of tasks / task-specific teachers / domains \\
        $\mathcal{D}_\tea^i$ & The $i$-th task-specific teacher's dataset \\
        $\phi_\tea^i$ & The $i$-th task-specific teacher's backbone \\
        & ($i=0$: the generalist's backbone)\\
        $h_\tea^i$ & The $i$-th task-specific teacher's head (classifier layer) \\
        & ($i=0$: the generalist's head) \\
        $\mathcal{D}_\proxy$ & The large unlabeled proxy dataset used for \\
        & distillation / consolidation \\
        $\phi_\stu$ & The distilled/consolidated student's backbone \\
        $h_\stu$ & The distilled/consolidated student's head (classifier layer) \\
        \bottomrule
    \end{tabular}
    }
    \resizebox{0.49\columnwidth}{!}{
    \begin{tabular}{cl} \toprule
        Symbol & Definition \\ \midrule
        $\mathcal{D}_\down^j$ & The $j$-th downstream task's dataset \\
        $\phi_\down^j$ & The consolidated student backbone fine-tuned on dataset $\mathcal{D}_\down^j$ \\
        $h_\down^j$ & The downstream model's head (classifier layer \\
        & or linear SVM) for dataset $\mathcal{D}_\down^j$ \\
        $\lambda_i$ & Loss weight for the $i$-th teacher \\
        t-split & First 50\% random split of classes of a dataset, used as $\mathcal{D}_\tea^i$ \\
        d-split & Second 50\% random split of classes of a dataset, used as $\mathcal{D}_\down^j$ \\
        & \\
        & \\
        & \\
        \bottomrule
    \end{tabular}
    }
    }
    \caption{Glossary of symbols.}
    \label{tab:glossary}
\end{table}

\section{Representation Consolidation}
\label{sec:method}

\paragraph{Problem statement.} We start with a collection of one or more task-specific image classification models $\{\mathcal {M}_\tea^i\}_{i=1}^N$, trained on corresponding datasets $\{\mathcal{D}_\tea^i\}$ belonging to some domain (\eg satellite images, images of flowers, \etc). We assume models consist of a feature extractor or backbone $\phi_\tea^{i}(\cdot)$, composed with a classifier head $h_\tea^{i}(\cdot)$ so that $\mathcal {M}_\tea^i$ =  $h_\tea^{i}(\phi_\tea^i(\cdot))$. We first consolidate the knowledge of these task-specific teachers into a single student representation $\phi_\stu(\cdot)$ using a proxy dataset $\mathcal{D}_\proxy$ (\eg, ImageNet)
and then fine-tune the student representation on a given downstream $\mathcal{D}_\down^{j}$ chosen from some set $\{\mathcal{D}_\down^{j}\}$. Our goal is that the resulting downstream model $h_\down^{j} (\phi_\down^j(\cdot))$ achieves good performance, where $\phi_\down^j$ denotes the student representation $\phi_\stu$ after tuning on $\mathcal{D}_\down^{j}$.
Figure~\ref{fig:method_illust} highlights how this differs from standard distillation in which the student model $h_\stu(\phi_\stu(\cdot))$ is simply evaluated on the same task its teachers were once trained to perform.


\paragraph{Forgetting and representation collapse during distillation.} 
We observe (Sec. ~\ref{sec:exp_motivation}) that standard knowledge distillation is insufficient to produce good student representations $\phi_\stu$. Simple distillation yields models that under-perform general pre-training representations (\eg with ImageNet) when evaluated on downstream tasks from the teachers' domain, and drastically under-perform on tasks outside the teachers domain. This holds true even though the student network is itself initialized with a general pre-trained representation. We argue that distillation from task-specific teachers thus suffers from catastrophic forgetting of general knowledge that is crucial for transfer learning.

Intuitively, transfer performance depends how distinguishable different classes are represented in the penultimate layer feature space.
When using traditional distillation, the student only learns from task-specific teachers trained on smaller datasets and is only required to discriminate the classes in those datasets. This is well suited for traditional distillation that evaluates on these same tasks. But for representation consolidation, distinguishing \emph{unknown downstream} classes requires preserving general features that may not be relevant to the teachers' specific task. Our strategy is thus to ensure that the student maintains general features as it learns task-specific features.



\paragraph{Method.} We use multi-head multi-task distillation and avoid forgetting with the help of a generalist teacher. During distillation we use $N+1$ heads, $h_{\stu}^0,\dots,h_{\stu}^N$, on top of the student backbone $\phi_\stu(\cdot)$.  In addition to 
the set of task-specific teachers $h_{\tea}^i(\phi_{\tea}^i(\cdot))$, $i\in\{1,\dots,N\}$, we also include a generalist teacher that was trained on ImageNet (denoted as $h_{\tea}^0(\phi_{\tea}^0(\cdot))$) and optimize student parameters using the loss:
\begin{equation}
    \mathcal{L} = \sum_{x \in \mathcal{D}_\proxy}\sum_{i=0}^N \lambda_i \mathcal{L}_{\text{distill}} \left( h_{\tea}^i(\phi_{\tea}^i(x)),~ h_{\stu}^i(\phi_{\stu}(x)) \right)
    \label{eq:loss}
\end{equation}
where $\mathcal{L}_{\text{distill}}$ is the distillation loss~\cite{Hinton2015DistillingTK}, \ie cross-entropy with temperature $T=2$:
\begin{equation}
    \mathcal{L}_{\text{distill}}(p_\tea, p_\stu) = -\sum_{c=1}^C p_\tea^{(c)} \log (p_\stu^{(c)})
\end{equation}
where $c$ indexes the $C$ classes, and $p_\tea = \text{softmax}\left(h_{\tea}^i(\phi_{\tea}^i(x)) / T \right)$, $p_\stu = \text{softmax}\left(h_{\stu}^i(\phi_{\stu}^i(x))/T\right)$.

We initialize the student backbone $\phi_{\stu}$ and its 0-th head $h_{\stu}^0$ using the generalist model's weights $\phi_\tea^0$, $h_\tea^0$, whereas other heads are randomly initialized. Since it is important to maintain pre-trained model's representational power, we simply set the loss weights $\lambda_0=1$ and $\lambda_i=\frac{1}{N}$ for $1 \leq i \leq N$. In this way, the learned representation $\phi_{\stu}$ must include features useful for both the pre-trained model's classes and other task-specific teachers' classes, improving suitability for future downstream transfer. After training the student we evaluate the resulting representation $\phi_{\stu}(\cdot)$ on multiple downstream tasks $\{\mathcal{D}_\down^{i}\}$.  For each task $j$ we can either fine-tune the whole model $h_\down^{j} (\phi_\down^j(\cdot))$ or keep the student representation fixed and only learn the classifier head $h_\down^{j} (\phi_\stu(\cdot))$ which is often referred
to as a {\em linear probe}. 




\section{Experimental setup}
\label{sec:exp_setup}
\paragraph{Datasets and downstream tasks.} We utilize datasets from a variety of domains to generate teachers and downstream tasks: Cars196~\cite{Krause20133DOR}, Resisc45~\cite{Cheng2017RemoteSI} (remote sensing images), iFood~\cite{Kaur2019FoodX251AD} and Food101~\cite{Bossard2014Food101M}, iFashion~\cite{Guo2019TheIF}, DTD~\cite{Cimpoi2014DescribingTI} (describable textures), iNaturalist~\cite{Horn2018TheIS} (species classification, 2019 challenge version), CUB Birds~\cite{Wah2011TheCB}, Flowers~\cite{Nilsback2008AutomatedFC}, Caltech256~\cite{Griffin2007Caltech256OC}, and Aircrafts~\cite{Maji2013FineGrainedVC}. Among these, iFood and Food101 are the same domain, and Birds and Flowers are subdomains of iNaturalist. We checked for near-duplicates between these datasets using perceptual hash~\cite{phash}, and found negligible duplication: 1 out of 134k iNaturalist (50\% classes we used) is a duplicate of CUB, and 8 out of 130k/100k images are duplicates between iFood and Food101.

To evaluate if a consolidated student representation has learned features relevant to a specific domain, we require downstream tasks that are similar to that of each task-specific teacher. Except for Food101 and iFashion, we split each dataset at random into two disjoint sets which each contain only 50\% of the classes. We take the first half of the dataset, named ``t-split'', and use as $\mathcal{D}_\tea^i$, to train a task-specific teacher. We use the second half of each dataset, named ``d-split'', as one of the downstream tasks $\mathcal{D}_\down^j$. When evaluating few-shot downstream transfer, we random sample 5 training images from each available class in $\mathcal{D}_\down^j$, but always use the entire test set for evaluation. We use all samples for those classes with $< 5$ images. For iFashion, which is a multi-task multi-label dataset, we use all images but with a subset of labels (\eg those related to clothing category) as $\mathcal{D}_\tea^i$ for teacher training. We then use all images with a disjoint set of labels (\eg those related to sleeve) as a downstream task $\mathcal{D}_\down^j$ for evaluating transfer. For iFashion's few-shot scenario, we randomly subsample 1000 images for downstream training, and evaluate on all test data.  We do not train any teacher on Food101 so the complete set of classes are used as a downstream task. 

For the large proxy dataset $\mathcal{D}_\proxy$ used  during distillation, we primarily use ImageNet~\cite{Russakovsky2015ImageNetLS} (ILSVRC12), but we also evaluate using Places365-standard~\cite{Zhou2018PlacesA1} or a concatenation of some of these datasets, which yield very similar results. 

\paragraph{Evaluation Criteria.} 
We note that our goal is not to produce state of the art representations for transfer learning, but rather to demonstrate an approach that reliably improves transfer performance of representations learned using distillation. We evaluate each student's $\phi_\stu$ using its performance when transferred to various downstream datasets $\{\mathcal{D}_\down^j\}$ and compare to multiple baselines for initializing the downstream representation $\phi_\down^j$:

(1) ImageNet-pretrained $\phi_\tea^0$ which is a strong baseline for transfer learning, (2) ImageNet-pretrained $\phi_\tea^0$ fine-tuned on the soft-labels produced by the ImageNet-pretrained model with batchnorms in test mode. This baseline aims to isolate the effect of soft-labels / self-distillation on model performance. (3) The task-specific teacher $\phi_\tea^1$  (or one of the teachers when $N>1$) without further distillation. (4) Traditional distillation with $N$ teachers, \ie without the ImageNet $h_\tea^0(\phi_\tea^0(\cdot))$ as a teacher. (5) Our consolidated representation $\phi_\stu$ which includes $h_\tea^0(\phi_\tea^0(\cdot))$ as a teacher.

We use the transfer accuracy on downstream tasks $\mathcal{D}_\down^j$ to measure each representation's power. We primarily use the linear probe (train a linear SVM as $h_{\down}^j$ over fixed $\phi$) on $\mathcal{D}_\down^j$'s training set (single training run for full dataset, few-shot performance averaged over 50 random subsampling trials). We also verify our results hold when fine-tuning the student representation $h_\down^j(\phi_\down(\cdot))$.

\paragraph{Implementation details of downstream training \& evaluation.} We will release our code to reproduce this paper upon publication. We use PyTorch.~\cite{NEURIPS2019_9015} For all network training, we use SGD with momentum of 0.9, weight decay of $10^{-4}$, a batch size of 32. Unless noted, we use a learning rate decay of $0.1$ at 50\% and 80\% of total training. We initialize the task-specific teacher training with a ResNet50~\cite{He2016DeepRL} pre-trained on ImageNet. We fine-tune each $\phi_\tea^i$ on the task-specific $\mathcal{D}_\tea^i$ for 120 epochs (learning rate decay at 70th and 100th epoch) while doing a log-scale grid search on the learning rate. For distilling $\phi_\stu$, our method is less sensitive to the choice of learning rate. We use a fixed learning rate of 0.01 for each $h_\stu^i$ and 0.001 for $\phi_\stu$, and a schedule of 40 epochs. Note that ImageNet pre-training in PyTorch uses 0.001 as the final epoch learning rate. This takes us roughly 4 days on an AWS  instance with an NVIDIA V100 for each of the 38 unique traditional distill or representation consolidation experiments. When the downstream transfer uses a fixed $\phi_\down^j$, we extract $\phi_\down(x)$ on the center image crop, and search $h_\down^j$'s SVM hyperparameters using a 5-fold cross-validation in scikit-learn~\cite{scikit-learn}. When the downstream transfer uses fine-tuning, we run a log-scale grid search of learning rate with a 50 epoch schedule.

\section{Results}
\label{sec:results}

\begin{table}[t]
\centering
\begin{subfigure}{\columnwidth}
\centering
\begin{tabular}{rc|ccc} \toprule
Dataset    & iFood       & iFood           & Food101        & Resisc45          \\
    &  (t-split)    &  (d-split)    &         &  (t-split)       \\
    &           & 5-shot     & 5-shot     & 5-shot     \\
evaluated head    & $h_\tea^1$ or $h_\stu^1$ & SVM $h_\down$ & SVM $h_\down$ & SVM $h_\down$ \\
 \midrule
ImageNet pretrain    & --        & 28.9         & 37.0         & 70.7         \\
Teacher (t-split) & 74.4      & 34.8         & 42.8         & 53.3         \\
Traditional distill  & 72.8      & 35.3         & 44.0         & 53.6         \\
Repr. consolid. (ours) & 68.6      & 38.8         & 47.2         & 69.5         \\ \bottomrule
\end{tabular}
\caption{iFood as task specific teacher dataset}
\end{subfigure}
\begin{subfigure}{\columnwidth}
\centering
\begin{tabular}{rc|ccc} \toprule
Dataset    & Resisc45      & Resisc45          & iFood         & Food101        \\
    & (t-split)     &  (d-split)        & (d-split)      &         \\
    &           & 5-shot     & 5-shot     & 5-shot     \\
evaluated head    & $h_\tea^1$ or $h_\stu^1$ & SVM $h_\down$ & SVM $h_\down$ & SVM $h_\down$ \\
    \midrule
ImageNet pretrain  & --          & 70.7         & 28.9         & 37.0         \\
Teacher (t-split)  & 98.2        & 67.9         & 14.8         & 17.9         \\
Traditional distill  & 97.9      & 61.6         & 9.9          & 12.1         \\
Repr. consolid. (ours) & 97.0      & 72.6         & 28.8         & 36.4         \\ \bottomrule
\end{tabular}
\caption{Resisc45 as task-specific teacher dataset}
\end{subfigure}

\caption{Accuracy comparison for baselines and representation consolidation, as judged with traditional criteria (first column, original network head's old task performance) and transfer learning criteria (last 3 columns, downstream task performance of linear SVM head trained on related and unrelated downstream datasets). Baselines work well for the original task, but underperform in transfer learning. Ours either outperforms or matches the best of all baselines in all transfer scenarios.}
\label{tab:td_vs_rd}
\end{table}

\paragraph{Motivational analysis: traditional distillation \emph{vs.} representation consolidation}
\label{sec:exp_motivation}
To highlight the difference between representation consolidation and traditional distillation, we use either iFood or Resisc45 (t-split) as $\mathcal{D}_\tea^1$ to train teachers, and run distillation/consolidation with $N=1$. We test on iFood (d-split), Food101 (full), and Resisc45 (d-split) as downstream tasks (t-split and d-split are disjoint 50\% classes of each dataset; see Section~\ref{sec:exp_setup})

We compare the ImageNet $\phi_\tea^0$, the task specific teacher $\phi_\tea^1$, traditionally distilled $\phi_\stu$ (\ie only use $h_\tea^1(\phi_\tea^1(\cdot))$ as teacher), and our consolidated $\phi_\stu$ (\ie use $h_\tea^0(\phi_\tea^0(\cdot))$ and $h_\tea^1(\phi_\tea^1(\cdot))$ as teachers).
On downstream tasks $\mathcal{D}_\down^j$, we follow representation learning's evaluation protocol (linear probe): train linear SVM $h_\down$ on $\phi(x),~x\in\mathcal{D}_\down^j$, evaluate on $\mathcal{D}_\down^j$'s test set. The results are shown in Table~\ref{tab:td_vs_rd}.
We also evaluate each representation on the (upstream) teacher's task following the traditional distillation evaluation protocol, \ie directly evaluate $h_\tea^1(\phi_\tea^1(\cdot))$ or $h_\stu^1(\phi_\stu(\cdot))$ performance on the upstream task $\mathcal{D}_\tea^1$'s test set. 

If we only focus on the upstream task $\mathcal{D}_\tea^1$, then traditional (proxy) distillation almost matches the teacher's performance, and representation consolidation performs worse than both of them. However, if we instead focus on the downstream transfer performance on $\mathcal{D}_\down^j$, we see the opposite trend. In the case of the Food101 downstream task, we see that the representation power of both the generalist teacher (Imagenet) as well as the task specific teacher is lower than our representation consolidation. For the teacher-related tasks (i.e., downstream task which is disjoint classes split from the same dataset as $\mathcal{D}_\tea^1$) we see a benefit of consolidation over generic pre-trained features. On the unrelated downstream task (Resisc45), where the generalist teacher excels, we notice that our method almost matches the performance even though it includes an improved representation for the food task. We see a similar trend for Resisc45, where consolidated student outperform the teacher and the generalist on teacher-related tasks but still retains the performance of the generalist teacher on unrelated tasks. This clearly demonstrates the significant difference between upstream and downstream transfer.

\begin{figure}
    \centering
    \begin{subfigure}{0.55\columnwidth}
        \includegraphics[width=\columnwidth]{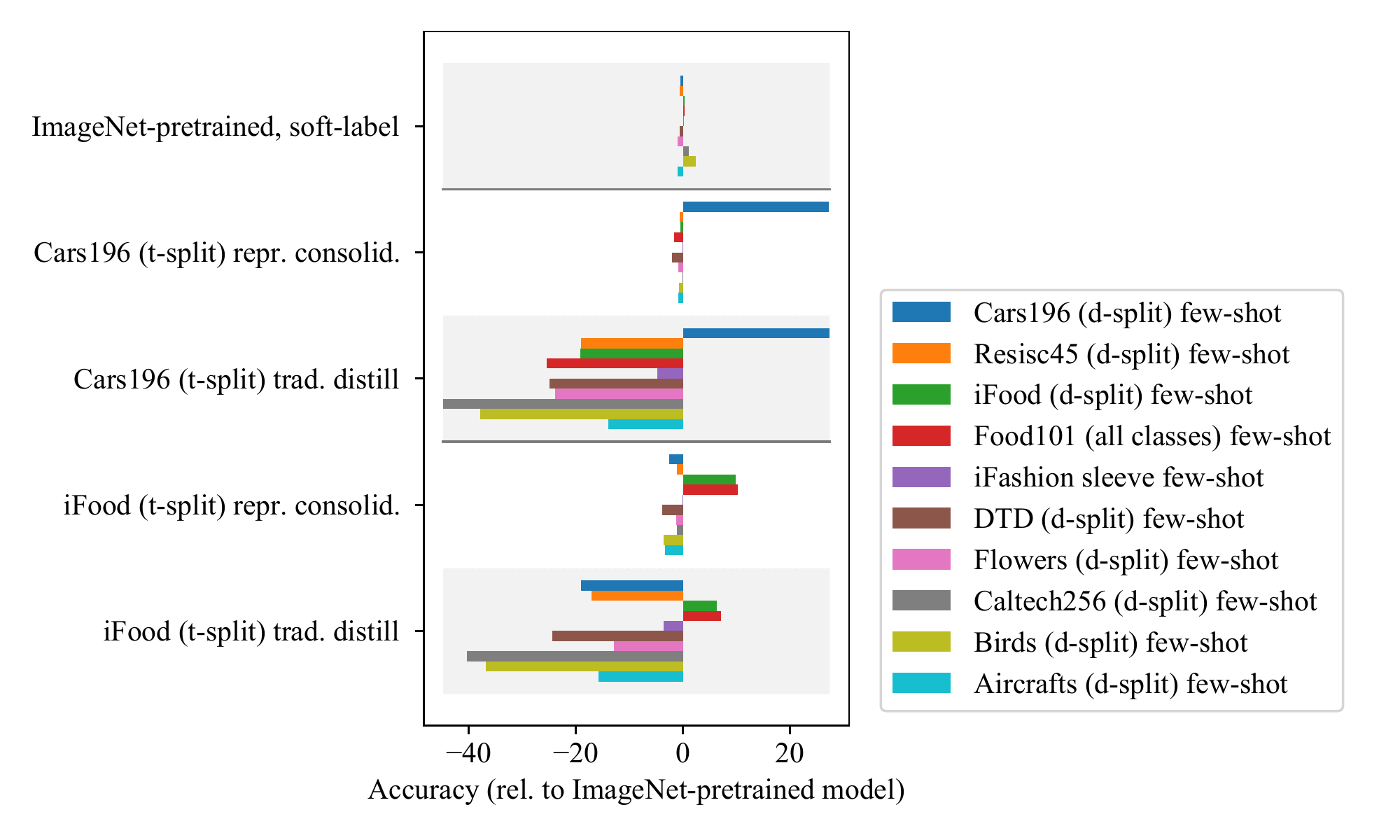}
        \caption{Excerpt (2 out of 10 $\mathcal{D}_\tea^1$ scenarios). See supplemental for full figure.}
        \label{fig:fix_1tea_plot}
    \end{subfigure}
    \begin{subfigure}{0.44\columnwidth}
\resizebox{\columnwidth}{!}{
\begin{tabular}{lcccccc} \toprule
              & \multicolumn{3}{c}{related $\mathcal{D}_\down^j$}         & \multicolumn{3}{c}{unrelated $\mathcal{D}_\down^j$}       \\[0.2em] \cmidrule(lr){2-4} \cmidrule(lr){5-7}
            & $>$ & $\approx$ & $<$ & $>$ & $\approx$ & $<$ \\ \midrule
our $\phi_\stu$ \vs ImageNet $\phi_\tea^0$   & 6          & 4       & 0            & 0          & 10      & 0            \\
our $\phi_\stu$ \vs trad. distill $\phi_\stu$ & 7          & 1       & 2            & 10         & 0       & 0            \\
trad. distill $\phi_\stu$ \vs ImageNet $\phi_\tea^0$   & 5          & 1       & 4            & 0          & 0       & 10          \\ \bottomrule
\end{tabular}
}
    \caption{Tally of comparisons (representation consolidation (ours) \vs traditional distill \vs ImageNet pre-trained model) among all 10 teacher dataset $\mathcal{D}_\tea^1$ scenarios.}
    \label{tab:fix_1tea_tally}
    \end{subfigure}
    \caption{\textbf{Left:} 5-shot linear SVM (fixed backbone) downstream transfer accuracy relative to ImageNet baseline. Comparing different representations with one task-specific teacher (y-axis) on a variety of downstream datasets (see legend). Representation consolidation (ours) in clear background. Excerpt (2 out of 10 domains). \textbf{Right:} Tally of comparisons among all ten $\mathcal{D}_\tea^1$ domains. On downstream tasks related to the teacher's, we outperform or match ImageNet, and on other tasks we match ImageNet performance. Traditional distill often underperforms ours (7/10 related, 10/10 unrelated) and ImageNet (4/10 related, 10/10 unrelated). See supplemental for full results.}
    \label{fig:fix_1tea}
\end{figure}

\paragraph{Improving student representation when $N=1$.} We show that this advantage of consolidation over baselines holds for a wide range of upstream and downstream datasets. Figure~\ref{fig:fix_1tea} summarizes few-shot SVM accuracy using different representations relative to ImageNet-pretrained (See supplemental for the full figure and raw numbers).

The conclusions are similar -- consolidation outperforms (Cars196, Resisc45, iFood, CUB, Aircrafts, iNaturalist) or matches (iFashion, DTD, Flowers, Caltech256) ImageNet pre-trained model performance on related downstream tasks, and matches its performance on unrelated ones. In contrast, traditional distillation (1) underperforms our method on related downstream tasks for all teachers except Cars196, iFashion, and Aircraft, and (2) drastically underperforms both consolidated and ImageNet features on unrelated downstream tasks. Notably, traditional distillation underperforms ImageNet even on related downstream tasks for Resisc45, DTD, Flowers, and Caltech256 teachers.

\begin{figure}[ht]
    \centering
    \begin{subfigure}{0.48\columnwidth}
        \includegraphics[width=\columnwidth]{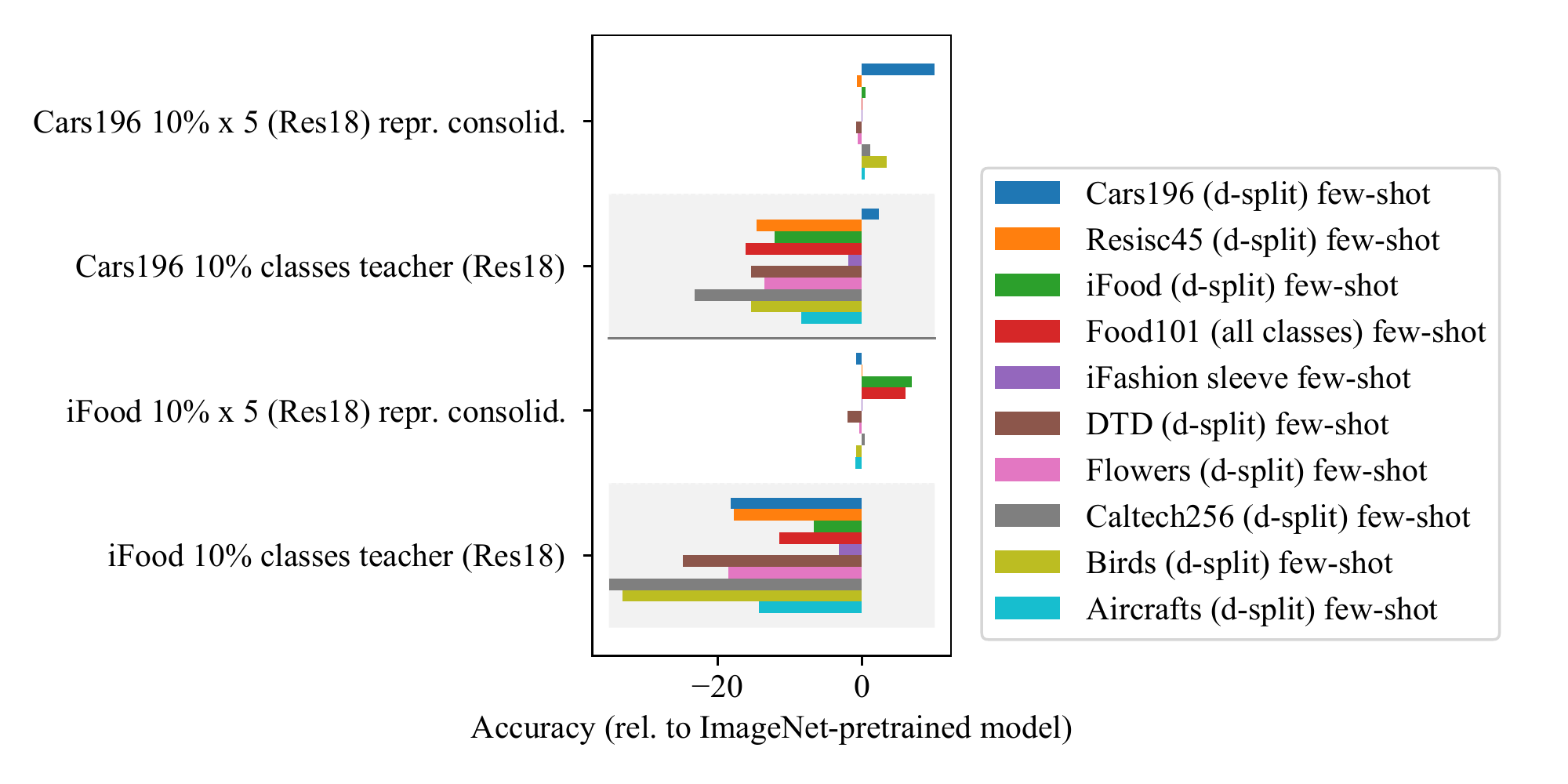}
        \caption{Merging same-domain ResNet18 teachers}
        \label{fig:fix_Ntea_r18}
    \end{subfigure}
    \begin{subfigure}{0.48\columnwidth}
        \includegraphics[width=\columnwidth]{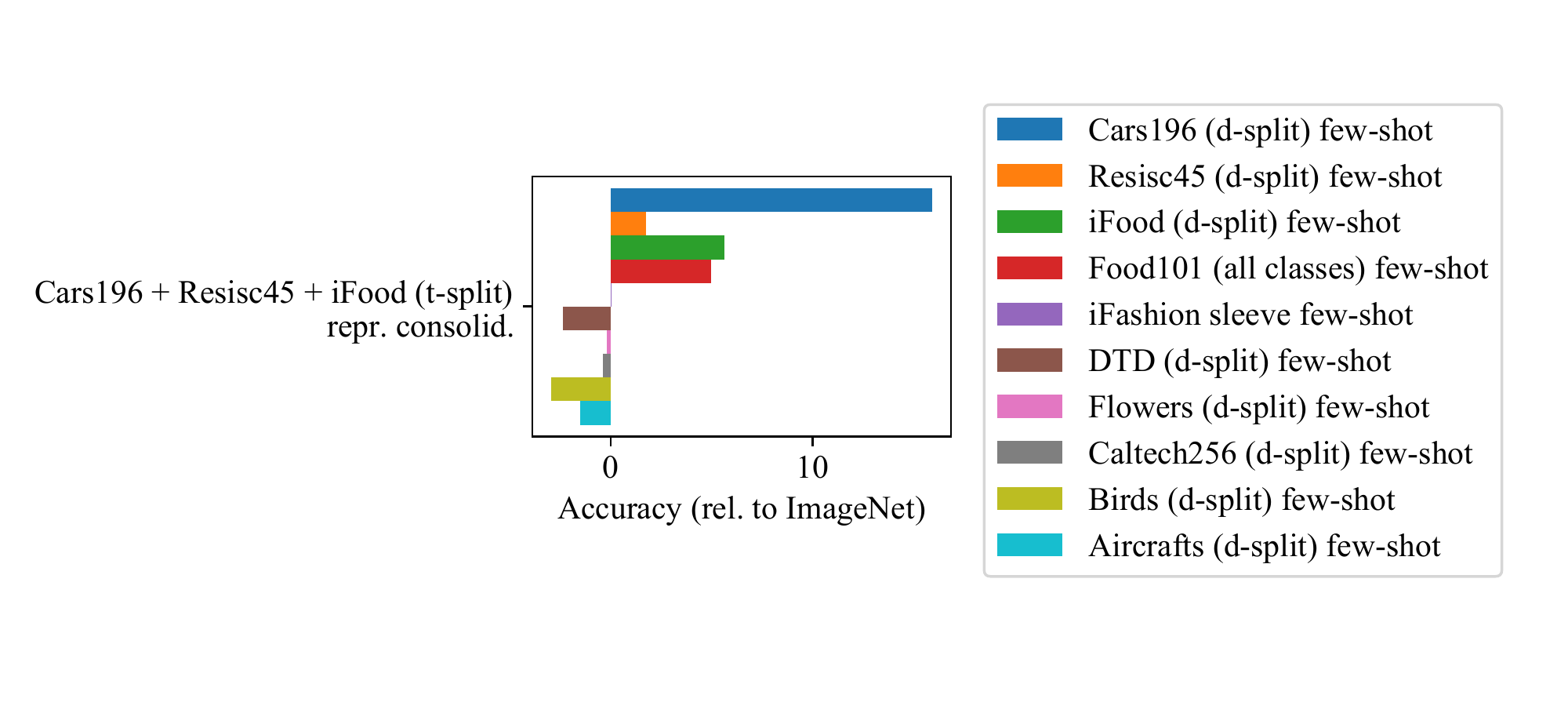}
        \caption{Merging multi-domain ResNet50 teachers}
        \label{fig:fix_Ntea_r50}
    \end{subfigure}
    \caption{Multi-model merging experiment. 5-shot linear SVM (fixed backbone) downstream transfer performance relative to ImageNet baseline. Comparing different merged multi-teacher representations (y-axis) on a variety of downstream datasets (see legend). Representation consolidation (ours) in clear background. (a) In the same domain, we are able to consolidate from models with different architectures and improve transfer performance over every single teacher. (b) We can consolidate different domain models and improve over the ImageNet representation. 
    }
    \label{fig:fix_Ntea}
\end{figure}

\paragraph{Consolidating representations with $N>1$.}  We can merge multiple task-specific teachers that are related to get a better representation. In addition, we show that our method is not constrained to merging models with the same architecture or $\phi_\tea^i$'s feature space dimensions like prior work~\cite{Geyer2019TransferLB}. To illustrate this, we split the ``t-split''  of Cars196 and iFood into five equal splits containing 10\% of the original classes. We train a ResNet18 teacher on each of the five splits. Then, we use either traditional distillation or representation consolidation to merge these models with a ResNet50 generalist teacher. We compare the resulting representations with the teacher model's and ImageNet. Figure~\ref{fig:fix_Ntea_r18} shows this result. For downstream tasks, we obtain better performance than using only one of the five teachers on both related and unrelated tasks, especially for iFood and Food101 where the teachers themselves underperform on related downstream task. This shows that our method can benefit from even teachers whose representation is weaker, as long as they have domain knowledge. 

Full comparison with traditional distillation and representation consolidation from only one of the five teachers are in supplemental material -- we gain performance on similar $\mathcal{D}_\down^j$ by using five task-specific teachers instead of one, and we outperform traditional distillation on related downstream tasks for iFood. We also show results when using all ResNet50 teachers for the five splits in the supplemental material. The results are similar and the conclusions are identical.


Finally, we explore merging models from different domains to form a multi-domain consolidated student. See Figure~\ref{fig:fix_Ntea_r50}. We observe that we can outperform ImageNet on all related downstream tasks, but the performance gain is smaller than representation consolidation in just one domain. We show in the supplemental material that we outperform traditional distillation on most downstream tasks except Cars196. 

\begin{figure}
    \centering
    \begin{subfigure}{0.55\columnwidth}
        \includegraphics[width=\columnwidth]{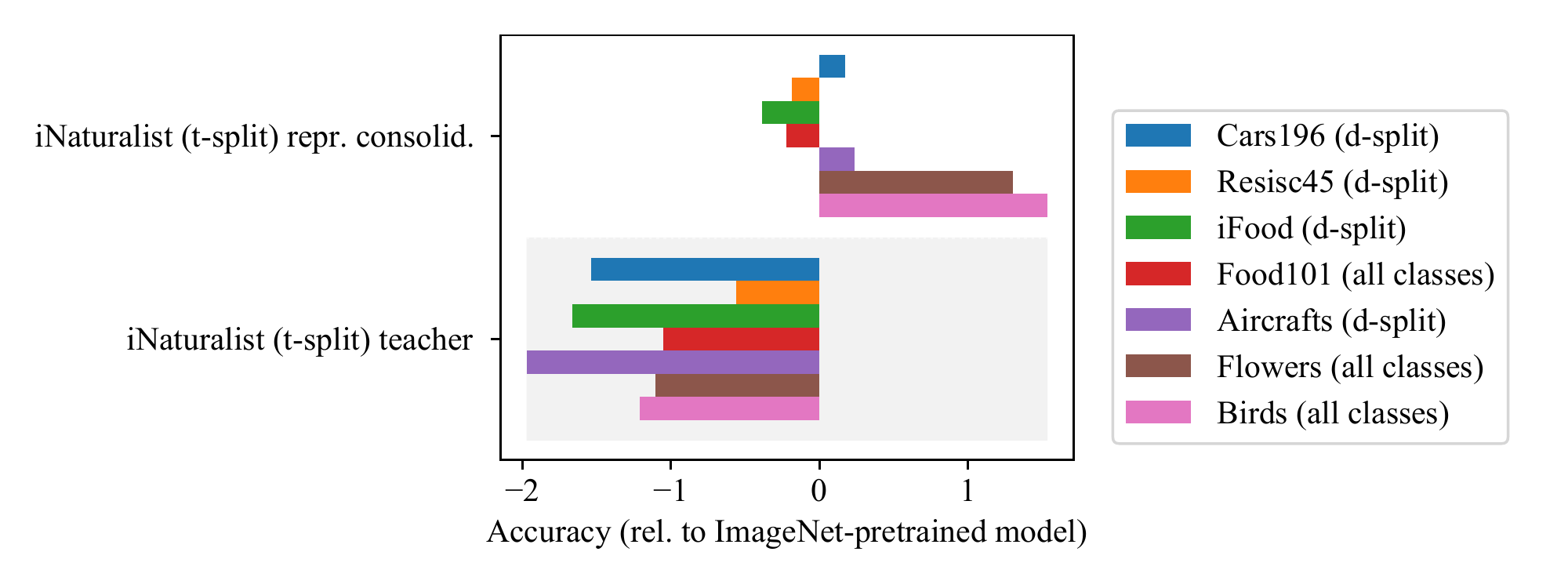}
        \caption{Excerpt of full-shot fine-tuning results (full dataset transfer for Flowers and Birds)}
    \end{subfigure}
    \caption{Fine-tuning downstream transfer performance relative to ImageNet baseline. Comparing ours (clear background) \vs traditional distill representations on a variety of downstream datasets (see legend). Our conclusions with the fine-tuning scenario is the same as the fixed representation scenario in Figure~\ref{fig:fix_1tea}. See supplemental material for the full results (few-shot, full-shot, more datasets).}
    \label{fig:ft_all}
\end{figure}

\paragraph{Fine-tuning downstream.} 
We also verify that our conclusions generalize to fine-tuning as well, especially without few-shot sampling. Figure~\ref{fig:ft_all} shows an excerpt of our results that include full dataset transfer. Fine-tuning allows the representation to change into one that better suits the downstream task, so the benefit of teachers' domain knowledge shrinks compared to fixed $\phi_\stu$ with linear SVM $h_\down^j$. Despite this, the conclusions are the same as the fixed $\phi_\stu$ scenario.  See supplemental material for few-shot fine-tuning and transfer with full d-splits, whose conclusions are the same as this section's.

\begin{figure}
    \centering
        \begin{subfigure}{0.48\columnwidth}
        \includegraphics[width=\columnwidth]{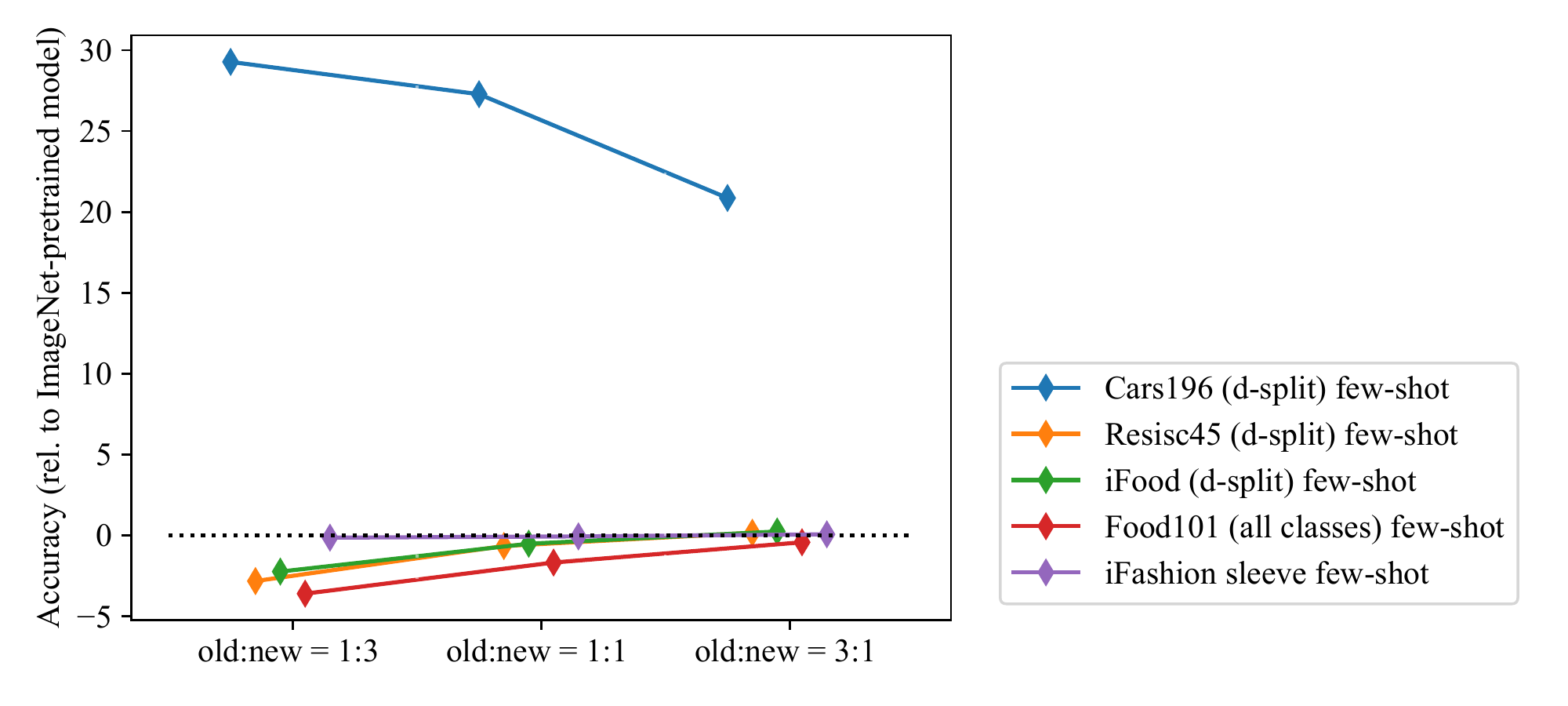}
        \caption{Tuning loss weights $\lambda_0$ (old) and $\lambda_1$ (new) result in different balance between related and unrelated downstream task performance. $\mathcal{D}_\tea^1=$Cars196, fix-representation linear probe.}
        \label{fig:misc_weights}
    \end{subfigure}\hspace{1em}
    \begin{subfigure}{0.48\columnwidth}
        \includegraphics[width=\columnwidth]{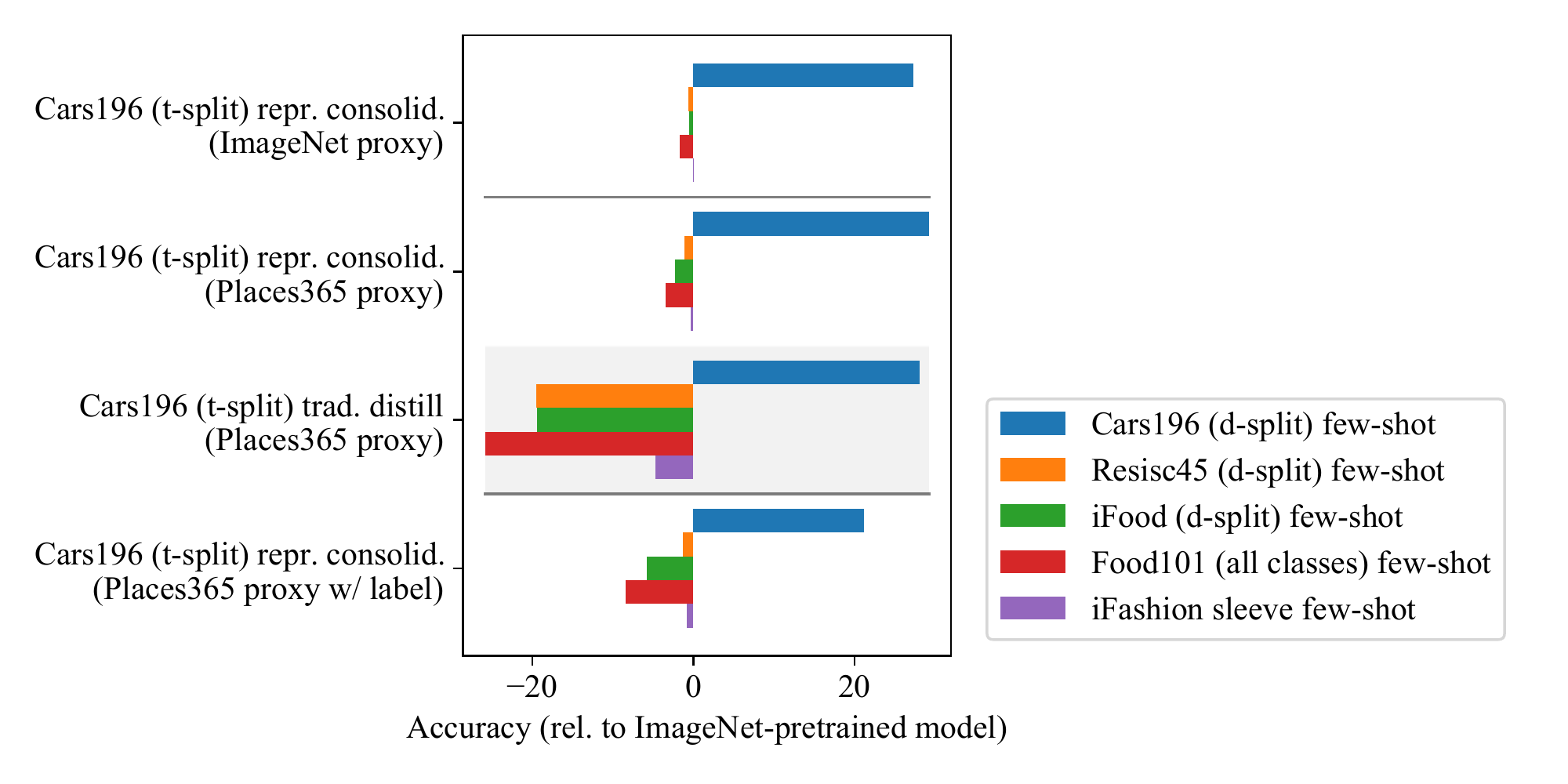}
        \caption{Using Places365 as $\mathcal{D}_\proxy$ has similar results as using ImageNet. However, training also on Places365's supervised labels hinders performance.}
        \label{fig:misc_places}
    \end{subfigure}
    \caption{\Misc studies. 5-shot linear SVM (fixed backbone) downstream transfer performance relative to ImageNet baseline. Representation consolidation (ours) in clear background. 
    }
    \label{fig:side}
\end{figure}
\paragraph{Influence of the Loss weights.} We show in Figure~\ref{fig:misc_weights} that our results are sensitive to the loss balance -- when we use $\lambda_0=\frac 3 2$, $\lambda_i=\frac 1 {2N}$ ($i>0$), we gain performance on the related downstream task but lose performance on unrelated ones, whereas using $\lambda_0=\frac 1 2$, $\lambda_i=\frac 3 {2N}$ ($i>0$) gives us the opposite. This allows us to trade-off the influence of each teacher, potentially suppressing unreliable ones.


\paragraph{Proxy data choice.} Figure~\ref{fig:misc_places} shows results replacing ImageNet with Places365 as the $\mathcal{D}_\proxy$ dataset. This yields similar performances as using ImageNet, only slightly weaker on unrelated downstream tasks. This shows representation consolidation is somewhat robust to the choice of proxy data. In the last row, we add a head to the student to learn the supervised ground truth of Places365 with a cross-entropy loss with loss weight of 1. This surprisingly hurt performance on both related and unrelated tasks. This shows that the effect of representation consolidation preserving general knowledge is largely due to the loss between $h_{\tea}^0$ and $h_{\stu}^0$, not because of any new information from input images.

\section{Related work}
\label{sec:related}
Our goal is to maximize the \emph{transferability} of a representation consolidated from multiple teachers. This downstream transfer aspect has received little attention in the literature (see e.g.,~\cite{Liu2019KnowledgeFI,Geyer2019TransferLB,Tian2020ContrastiveRD}), but this problem formulation is closely related to prior work on multi-model merging and distillation with proxy data.

\paragraph{Multi-model merging.} It is useful to combine separate models that perform different tasks into a single model for efficiency and performance benefits. 
Knowledge Concentration~\cite{Gao2017KnowledgeCL} combines teachers trained on subsets of 100k classes in the EFT dataset~\cite{Gao2017KnowledgeCL} into one single model that outperforms direct training on the full data, using handcrafted sparse connections for the final student layers.
Chou~\etal~\cite{Chou2018UnifyingAM} merge CNNs by combining the kernel weights of different models by clustering and lookup, and then fine-tuning the resulting weights.
Zhao~\etal~\cite{Zhao2020ObjectDW} merge object detection models when some classes may be the background to other models.
Vongkulbhisal~\etal~\cite{Vongkulbhisal2019UnifyingHC} combines models with different but potentially overlapping classes into one using the combined dataset, by computing unified soft labels using known intra-model class correspondence. 
Ye~\etal~\cite{Ye2020DataFreeKA} progressively train a GAN to regenerate proxy data for all teachers, and transfer all teachers into one student layer by layer. 
Chakraborty~\etal~\cite{Chakraborty2018AMM} aggregates an ensemble of weak classifiers by weighted average of their outputs.
Park~\etal~\cite{Park2020FeatureLevelEK} merges an ensemble for the same task using both the distillation loss and an adaptation layer to predict the teacher's penultimate layer's activations.
In one-shot federated learning~\cite{Guha2019OneShotFL}, multiple clients with their own private data train a model each while protecting their data from being shared or otherwise leaked, and merge the models at the end of the training using an ensemble or distillation.

Unlike our approach, these methods merge models in order to perform exactly the \emph{same task} as the teacher models, are not concerned with the performance of the student when transferred to a downstream task, and often~\cite{Gao2017KnowledgeCL,Chou2018UnifyingAM,Zhao2020ObjectDW,Vongkulbhisal2019UnifyingHC,Chakraborty2018AMM,Park2020FeatureLevelEK} require revisiting the original training images. We have show that these teachers themselves have poor transferrability compared to a simple pretrained baseline, and merging them yields suboptimal transfer learning performance.

\paragraph{Multi-model merging for transfer.}
Knowledge Flow~\cite{Liu2019KnowledgeFI} connects the student to multiple teachers' intermediate layers to kick-start its training, and gradually penalizes its reliance on teacher models over the training of the final target task. 
Geyer~\etal~\cite{Geyer2019TransferLB} uses IMM~\cite{Lee2017OvercomingCF} to merge multiple models using their diagonal approximated Fisher information matrix (FIM), balancing their importance using learned weights, to form a new representation to fine-tune from. Computing the FIM requires reprocessing the original teacher training data (unlike our approach that only needs generic proxy data).  Furthermore,  this method requires all students and teachers to have exactly the same backbone architecture while ours works on any combination of student and teacher architectures. These methods directly optimize performance on the downstream task and thus require separate model merging runs for each different target dataset. Our approach is more efficient, as it only requires consolidating teachers once to improve the pre-trained representation independently of the downstream task. Finally, we note that the representations learned were not compared to a strong baseline (i.e., pretraining on ImageNet), which we argue is a prerequisite for actually being useful in  real applications. One of our main contributions is observing the need for including a generalist teacher, an insight which is largely orthogonal to, and could be combined with these previous approaches.

\paragraph{Distillation with representation losses (``representation distillation'')} tries to capture additional structure of feature representations by aligning student and teacher feature activations during distillation. Koratana~\etal~\cite{Koratana2019LITLI} compress models by adding $L_2$ losses between intermediate representations of the teacher and students. Aguilar~\etal~\cite{Aguilar2020KnowledgeDF} use KL divergence and cosine similarity to make the attention and representation of intermediate features of the student and teacher similar. Tian~\etal~\cite{Tian2020ContrastiveRD} adds contrastive representation learning loss to the penultimate layer to preserve feature structures, by maximizing each image's student and teacher features' mutual information. They demonstrate this can yield better representations for transfer than traditional knowledge distillation but don't consider multiple downstream tasks or compare to strong generalist baselines.

\paragraph{Proxy data (``data-free'') distillation} transfers the input-output function of a teacher network to a student network without using the teacher's training data. \cite{Yalniz2019BillionscaleSL,Orekondy2019KnockoffNS} opt to use a large general proxy dataset to query the teacher, and their teacher outputs to on this data to train the student. Other methods ~\cite{Nayak2019ZeroShotKD,Chen2019DAFLDL,Haroush2020TheKW,
ChawlaDatafreeKD} generate proxy data directly from the trained models, and use this data to train the students. Further, ~\cite{Micaelli2019ZeroshotKT,Yin2020DreamingTD}, also encourage generating samples the student and teacher disagree on. Lastly, other methods require the original dataset to compute meta-data information such as feature cluster mean.~\cite{Lopes2017DataFreeKD,Bhardwaj2019DreamDA} Some train a GAN from the teachers to maximize chosen class predictions~\cite{Fang2019DataFreeAD,Yoo2019KnowledgeEW,Ye2020DataFreeKA}, sometimes also batchnorm statistics~\cite{Luo2020LargeScaleGD,Xu2020GenerativeLD}, and sometimes on proxy data instead.~\cite{Addepalli2020DeGAND,Besnier2020ThisDD} These works do not concern either the transferability of the learned student network  or merging multiple teachers.

\paragraph{Incremental learning} is also related to our overall goal of growing a library of expert representations. These methods continually learn tasks or classes but often with limited access to past training data~\cite{Li2018LearningWF,Rebuffi2017iCaRLIC,Kirkpatrick2017OvercomingCF,Zenke2017ContinualLT,Hu2019OvercomingCF,Yin2020DreamingTD,Prabhu2020GDumbAS}.
Our approach addresses many of the same challenges by consolidating knowledge in the form of feature representations without revisiting old data used to train teachers, but our ``increments'' are whole tasks whose labels don't necessarily overlap. 


\section{Summary}
In this paper, we show that traditional distillation can result in a representation suboptimal for downstream task transfer learning, because they only focus on preserving the end-to-end input-output mapping of the old task. We show our representation consolidation with the generalist model as an additional teacher preserves the transferability of the strong ImageNet baseline and improve the performance for both related and unrelated downstream tasks over traditionally distilled networks. We show that we can merge multiple models in the same domain to get a better representation than any single model.

\paragraph{Limitations and societal impact.} We assume we have perfect knowledge of which tasks form the same domain and which tasks do not belong to a domain. Our performance drops when teachers from different domains are consolidated. In future work, we plan to automatically determine how to cluster a large amount of teachers into domains. In addition, one of our contributions assume the existence of a strong representation baseline such as the ImageNet pre-trained model, which is true for \eg images and language, but not for other fields \eg 3D reconstruction. 
Our method also takes the teachers as is and learn from them, and any mistakes made by the teachers can be propagated during the consolidation. Possible mitigations include using a better teacher that makes less such mistakes, or using regularization on both teacher and student training, such as making similar inputs map to similar outputs.



{
\small
\bibliography{related_work,experiments}
\bibliographystyle{unsrt}



}

\storecounter{figure}{figurecounterstore}
\storecounter{table}{tablecounterstore}
\appendix

\section{Complete figures for Section~\ref{sec:results} experiments}

We now provide the full graphs and table for our experiments. Note that we have summarized these results and all conclusions in the main paper.

\begin{table}[b!]
    \centering
\resizebox{0.9\columnwidth}{!}{
\begin{tabular}{lcccccc} \toprule
              & \multicolumn{3}{c}{related $\mathcal{D}_\down^j$}         & \multicolumn{3}{c}{unrelated $\mathcal{D}_\down^j$}       \\[0.2em] \cmidrule(lr){2-4} \cmidrule(lr){5-7}
            & $>$ & $\approx$ & $<$ & $>$ & $\approx$ & $<$ \\ \midrule
our $\phi_\stu$ \vs ImageNet $\phi_\tea^0$   & (6 others)                                                                   & \begin{tabular}[c]{@{}c@{}}iFashion, \\ DTD, Flowers, \\Caltech256.\end{tabular} &                                                                                  &                                                                            & (all 10)                    &                                                                              \\ [2em]
our $\phi_\stu$ \vs trad. distill $\phi_\stu$ & (7 others)                                                                   & Cars196.                                                                          & iFashion, Aircraft.                                                               & (all 10)                                                                      &                          &                                                                              \\ [1em]
trad. distill $\phi_\stu$ \vs ImageNet $\phi_\tea^0$   & (5 others)                                                                   & iNaturalist.                                                                      & \begin{tabular}[c]{@{}c@{}}Resisc45, \\DTD, Flowers, \\ Caltech256.\end{tabular} &                                                                            &                          & (all 10)                                                                       \\
\bottomrule
\end{tabular}
}
    \caption{Detailed tally of Figure~\ref{sfig:fix_1tea} ($N=1$ single task-specific teacher) explaining Figure~\ref{tab:fix_1tea_tally}. 5-shot linear SVM (fixed backbone) downstream transfer. On downstream tasks related to the teacher's, we outperform or match ImageNet, and on other tasks we match ImageNet performance. Traditional distill often underperforms ours (7/10 related, 10/10 unrelated) and ImageNet (4/10 related, 10/10 unrelated).}
    \label{stab:fix_1tea_tally}
\end{table}
\begin{figure}
    \centering
    \includegraphics[width=\columnwidth]{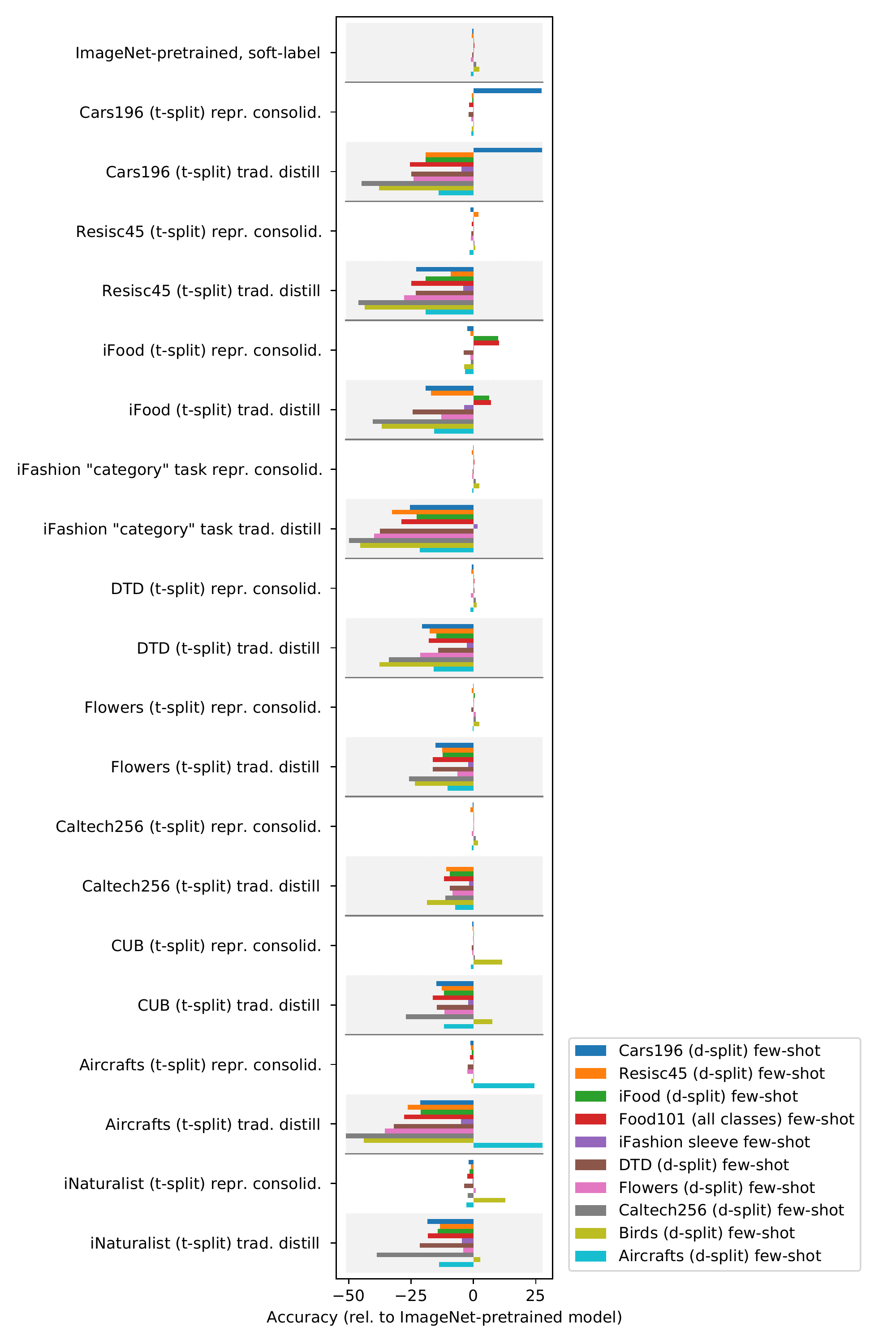}

    \caption{Full results for Figure~\ref{fig:fix_1tea_plot} ($N=1$ single task-specific teacher). 5-shot linear SVM (fixed backbone) downstream transfer accuracy relative to ImageNet baseline. Showing performance with each of the 10 $\mathcal{D}_\tea^i$ cases. This extensive set of experiments have the same conclusions as the main paper: we match or outperform ImageNet representation while traditional distill often underperforms. See also Table~\ref{stab:fix_1tea_tally} for a comparison tally.}
    \label{sfig:fix_1tea}
\end{figure}
\paragraph{Improving student representation when $N=1$.} Figure~\ref{sfig:fix_1tea} and Table~\ref{stab:fix_1tea_tally} show full results for Figure~\ref{fig:fix_1tea}. Please see the main paper for analysis and conclusions.

\begin{figure}
    \centering
        \includegraphics[width=\columnwidth]{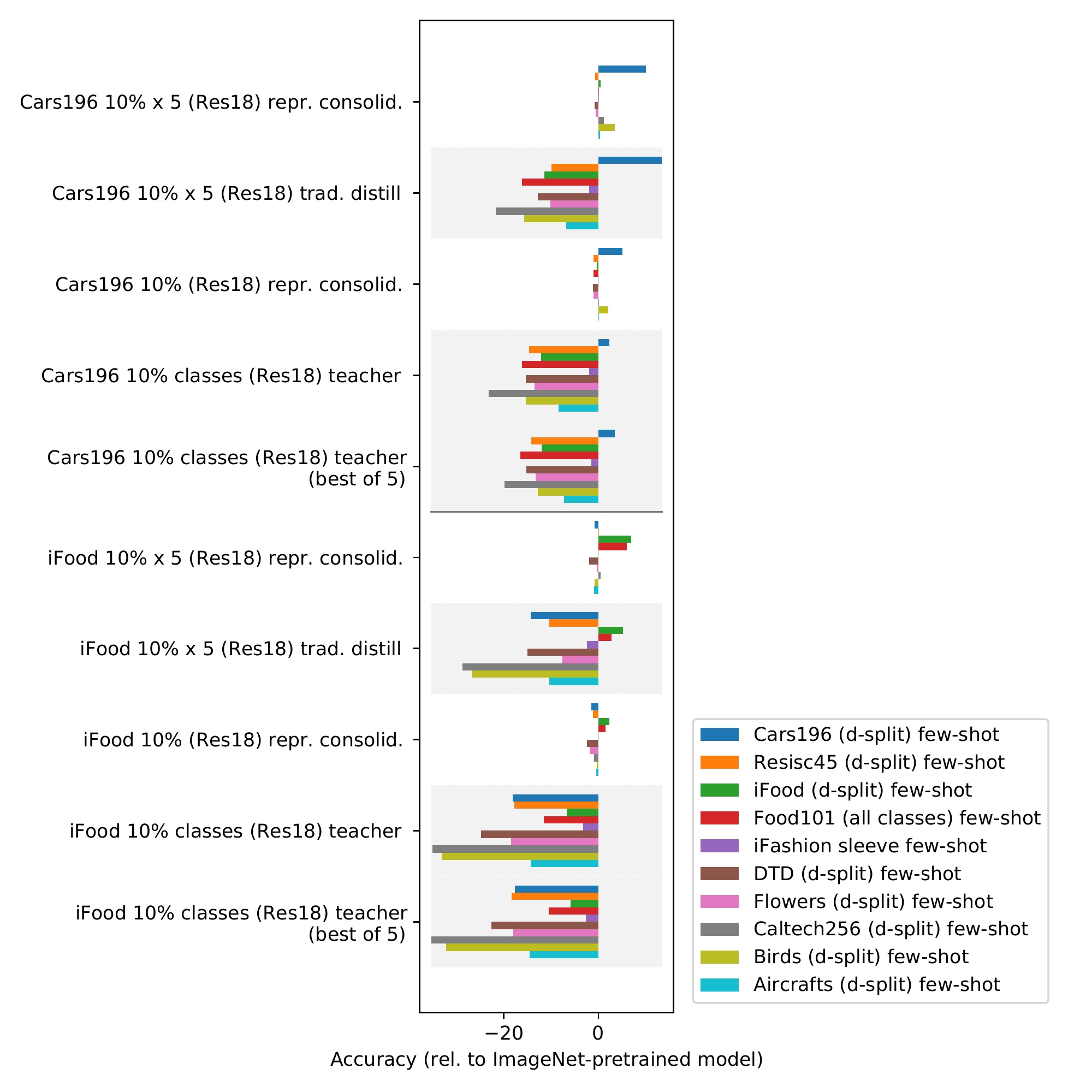}

    \caption{Merging same-domain ResNet18 teachers. Part 1/2 of full results for Figure~\ref{fig:fix_Ntea_r18} ($N>1$ multiple model merging, same-domain). 5-shot linear SVM (fixed backbone) downstream transfer. We are able to consolidate from models with different architectures (ResNet50 $\phi_\tea^0$ and ResNet18 $\phi_\tea^i$) and improve transfer performance over every single teacher. }
    \label{sfig:fix_Ntea_r18}
\end{figure}
\begin{figure}
    \centering
        \includegraphics[width=\columnwidth]{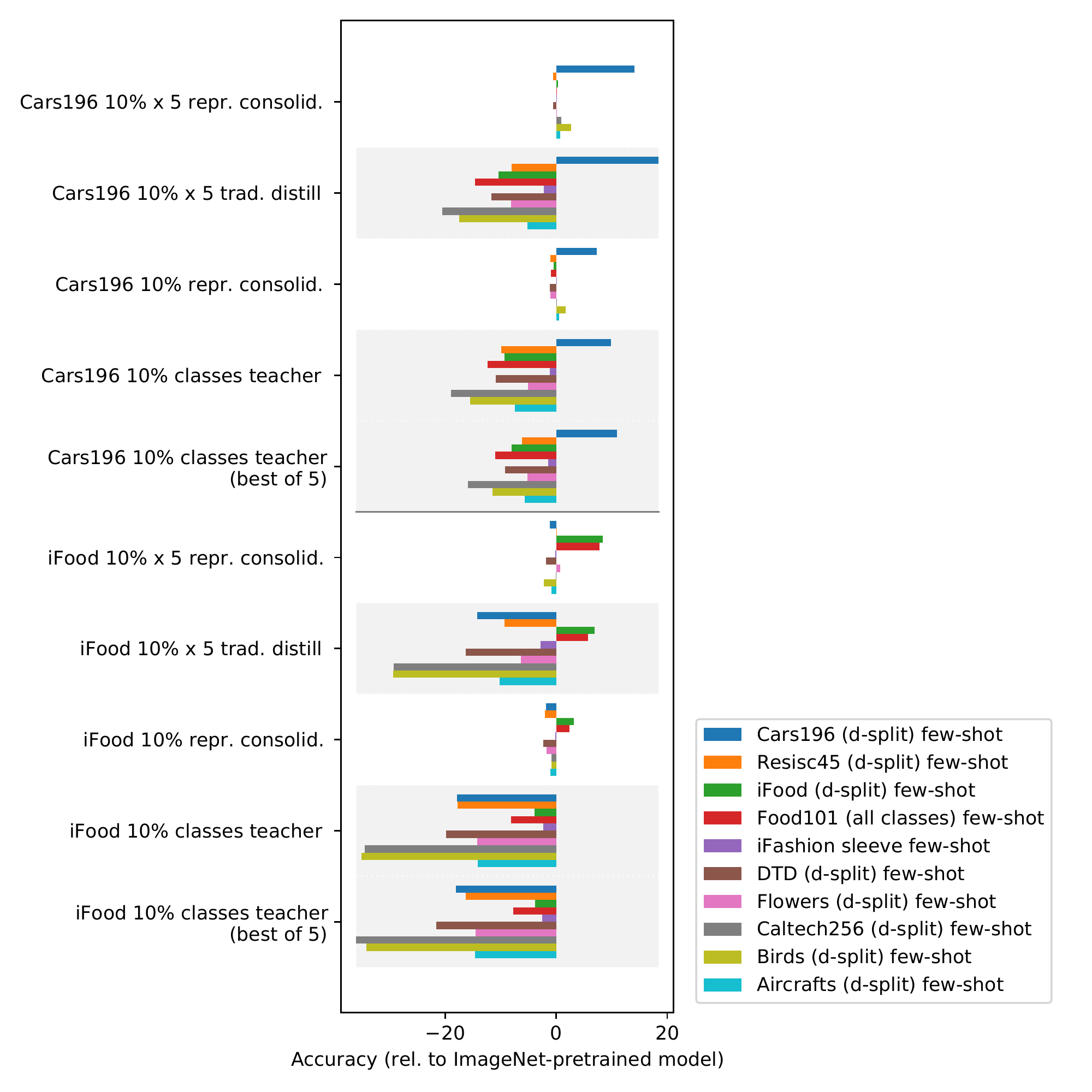}

    \caption{Merging same-domain ResNet50 teachers. Part 2/2 of full results for Figure~\ref{fig:fix_Ntea_r18} ($N>1$ multiple model merging, same-domain). 5-shot linear SVM (fixed backbone) downstream transfer. Full results (part 1/3) for Figure~\ref{fig:fix_Ntea} ($N>1$ multiple model merging). Our conclusions generalize to using all ResNet50 teachers.}
    \label{sfig:fix_Ntea_r50}
\end{figure}
\begin{figure}
    \centering
        \includegraphics[width=\columnwidth]{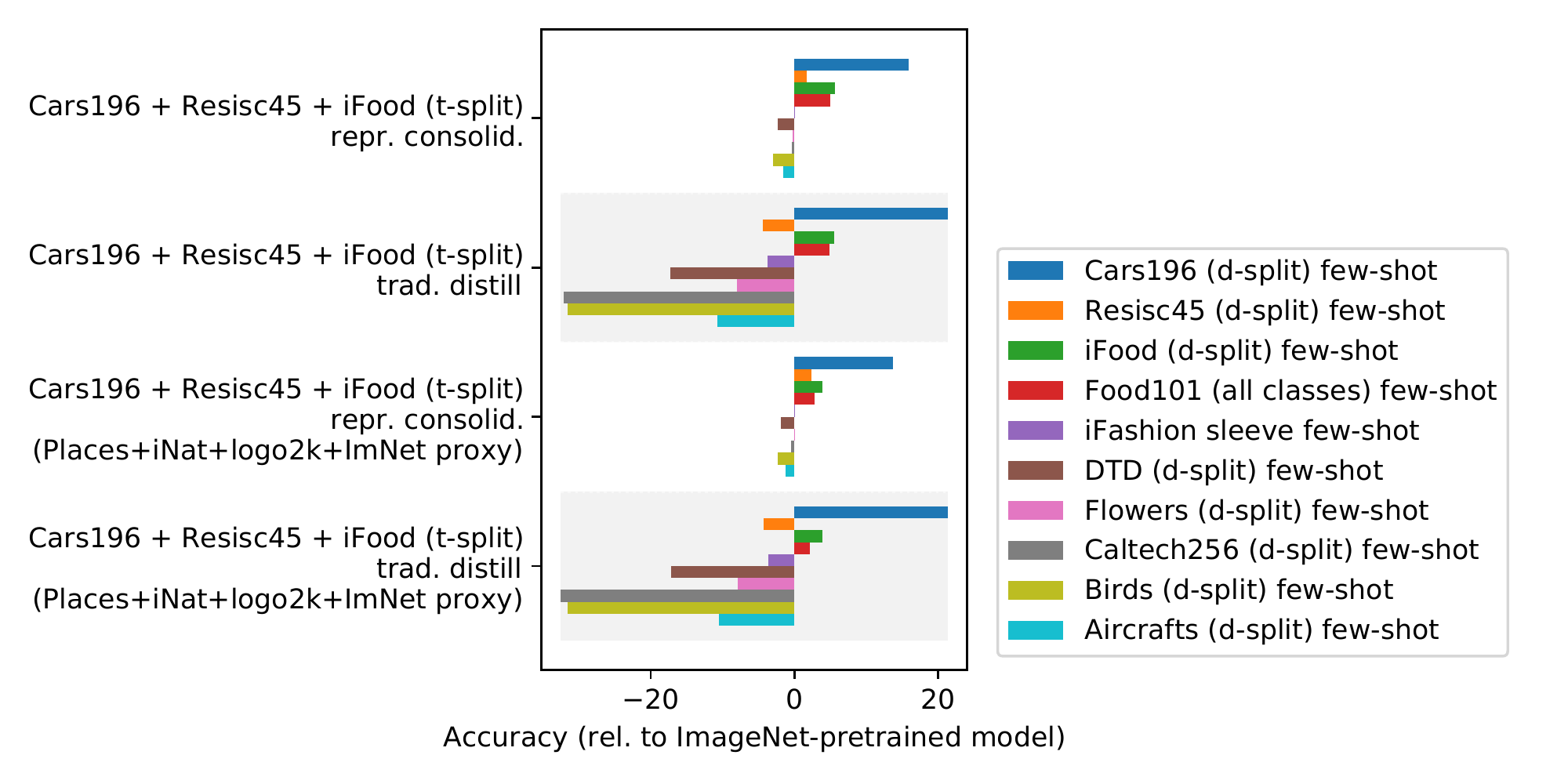}
    \caption{Merging different domains. Full results for Figure~\ref{fig:fix_Ntea_r50} ($N>1$ multiple model merging, multiple domains). 5-shot linear SVM (fixed backbone) downstream transfer. We can consolidate different domain models and improve over the ImageNet representation and (for most downstream datasets) multi-task traditional distillation.}
    \label{sfig:fix_Ntea_multidom}
\end{figure}
\paragraph{Consolidating representations with $N>1$.}  Figures~\ref{sfig:fix_Ntea_r18},~\ref{sfig:fix_Ntea_r50},~\ref{sfig:fix_Ntea_multidom} shows full results for Figure~\ref{fig:fix_Ntea}. 

For Figures~\ref{sfig:fix_Ntea_r18}~and~\ref{sfig:fix_Ntea_r50} same-domain model merging, in addition to the main paper results merging ResNet18 $\phi_\tea^i$ and ResNet50 $\phi_\tea^0$, we show similar results for all teachers being ResNet50 in Figure~\ref{sfig:fix_Ntea_r50}. We also show comparison to traditional distill with 5 teachers, representation consolidation with only 1 teacher, and the teacher itself (randomly chosen, or best out of 5 according to related $\mathcal{D}_\down^j$ performance). 
The conclusions are the same, and merging five teachers using representation consolidation outperforms all baselines on both related and unrelated downstream tasks (except Cars196 with related $\mathcal{D}_\down^j$ against traditional distill).

For multi-domain model merging in Figure~\ref{sfig:fix_Ntea_multidom}, we compare to traditional distill with multi-task learning. We outperform it on all related and unrelated $\mathcal{D}_\down^j$ except for Cars196. We also explore using a concatenation of multiple large unlabeled datasets as $\mathcal{D}_\proxy$. With a more diverse proxy, the performance drops a little for related $\mathcal{D}_\down^j$ and stays similar for unrelated $\mathcal{D}_\down^j$, suggesting we are somewhat insensitive to choice of datasets, but a more diverse proxy data may not provide better model merging performance.

\begin{figure}
    \centering
        \includegraphics[width=\columnwidth]{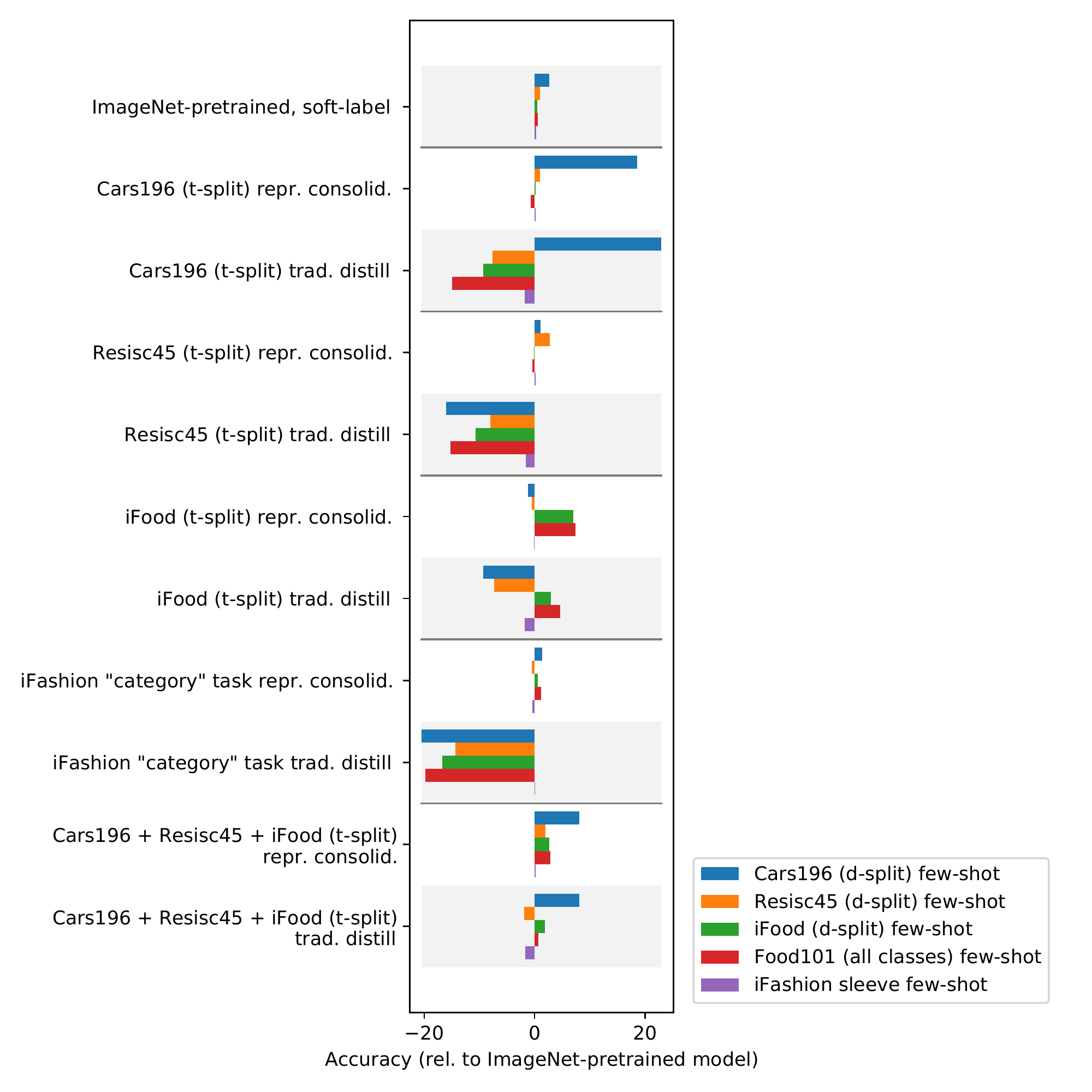}
    \caption{Few-shot fine-tuning results. $N=1$ for single domain as well as $N>1$ for multiple domains. Part 1/3 of full results for Figure~\ref{fig:ft_all} (fine-tuning downstream transfer). The same conclusions as the fixed representation scenario in Figures~\ref{fig:fix_1tea},~\ref{sfig:fix_1tea} hold.}
    \label{sfig:ft_fs}
\end{figure}
\begin{figure}
    \centering
        \includegraphics[width=\columnwidth]{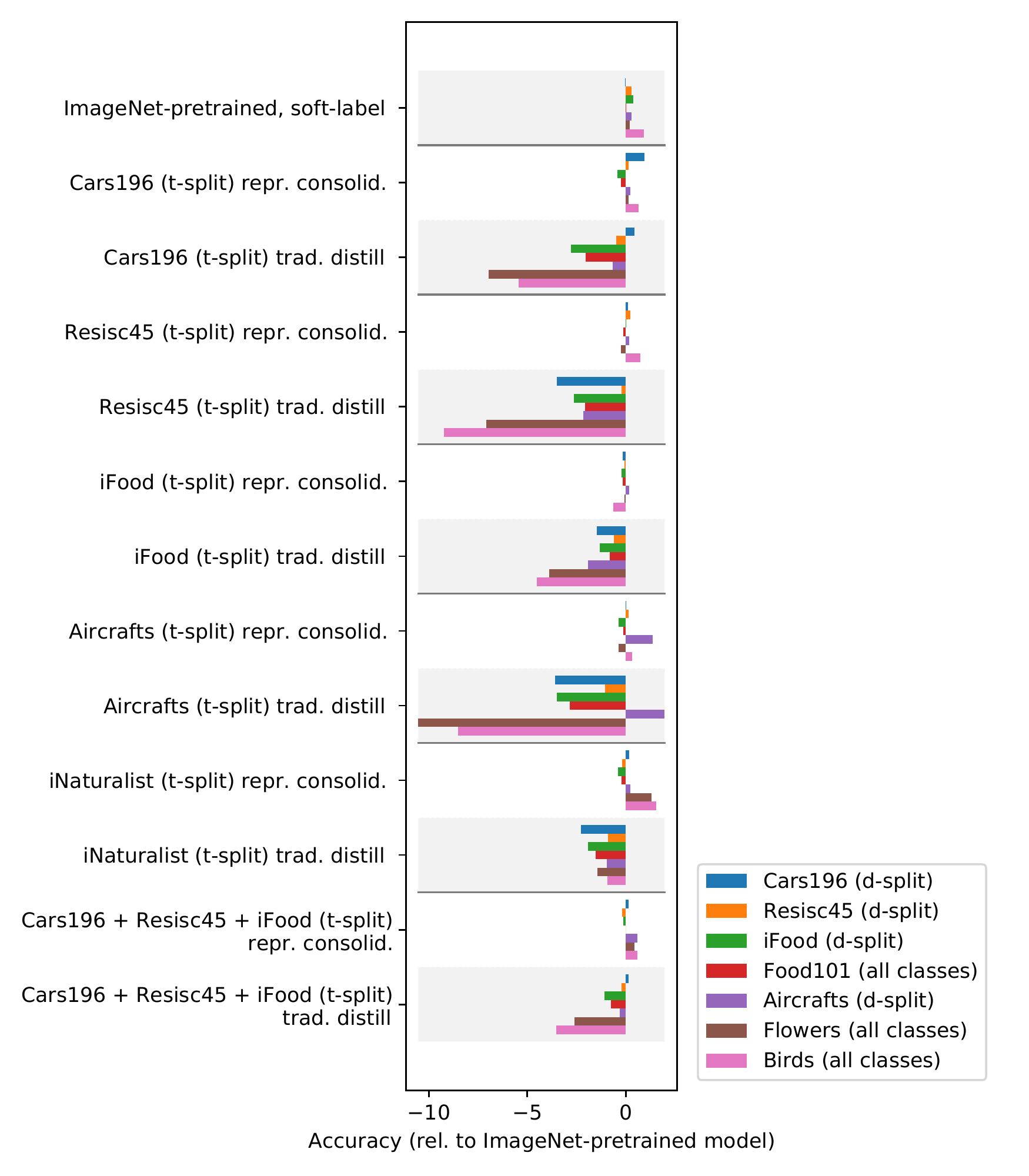}
    \caption{Full-shot fine-tuning results. $N=1$ for single domain as well as $N>1$ for multiple domains. Part 2/3 of full results for Figure~\ref{fig:ft_all} (fine-tuning downstream transfer). The same conclusions as the fixed representation scenario in Figures~\ref{fig:fix_1tea},~\ref{sfig:fix_1tea} hold.}
    \label{sfig:ft_ls}
\end{figure}
\begin{figure}
    \centering
        \includegraphics[width=\columnwidth]{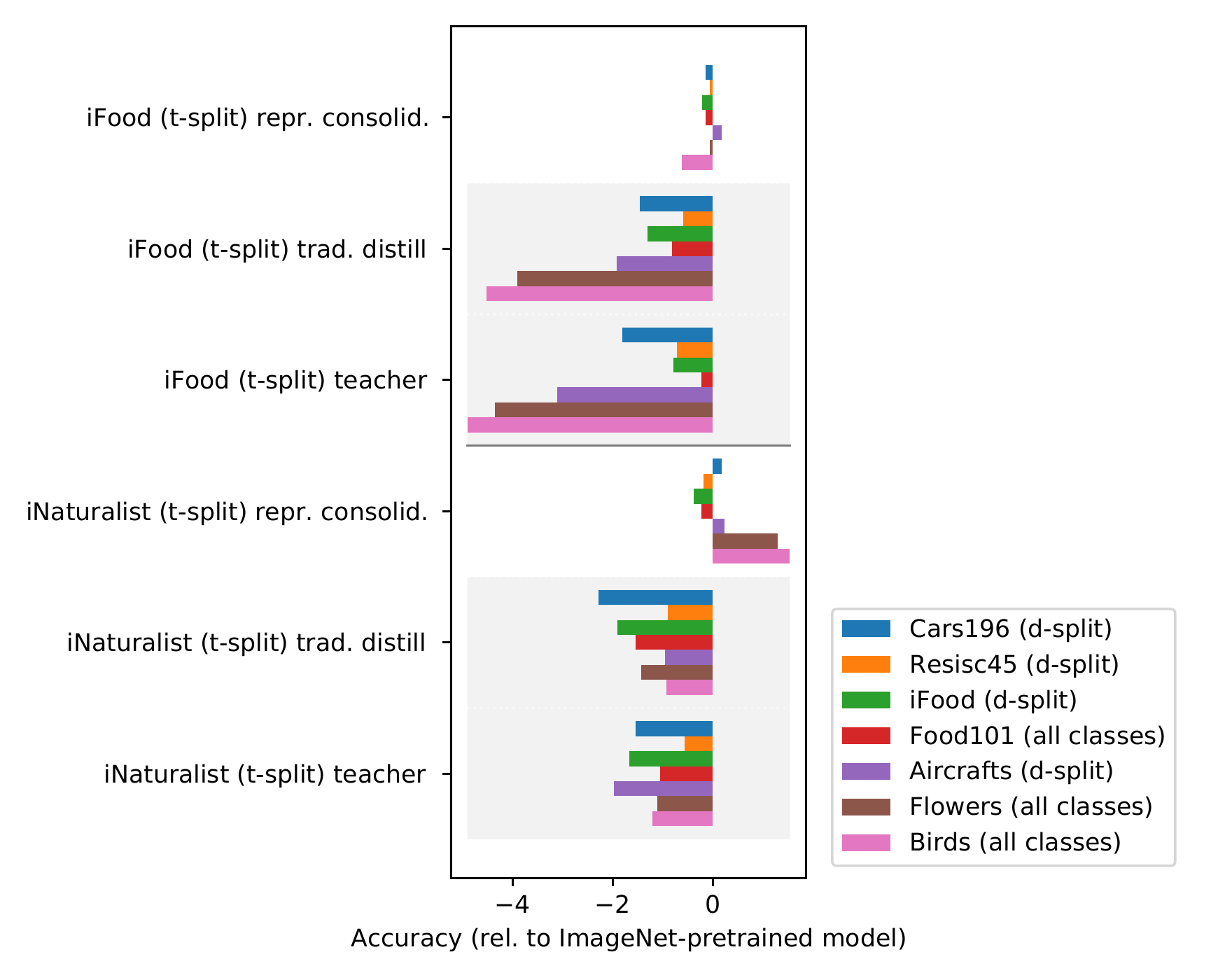}
    \caption{Full-shot fine-tuning results, also comparing to the teacher ($N=1$). Part 3/3 of full results for Figure~\ref{fig:ft_all} (fine-tuning downstream transfer). The same conclusions as the fixed representation scenario in Figures~\ref{fig:fix_1tea},~\ref{sfig:fix_1tea} hold.}
    \label{sfig:ft_ls_full}
\end{figure}
\paragraph{Fine-tuning downstream.} Figures~\ref{sfig:ft_fs},~\ref{sfig:ft_ls},~\ref{sfig:ft_ls_full} show full results for Figure~\ref{fig:ft_all}. These extensive results show that in both few-shot and full-shot scenarios, and both single-domain and multi-domain scenarios, we have the same conclusions as the fixed representation experiments.

\section{Raw number accuracy tables for all experiments} 
\begin{itemize}
    \item Table~\ref{stab:raw_fix} shows fixed representation few-shot results for Figures~\ref{fig:fix_1tea},~\ref{fig:fix_Ntea},~\ref{fig:side} in the main paper (Figures~\ref{sfig:fix_1tea},~\ref{sfig:fix_Ntea_r18},~\ref{sfig:fix_Ntea_r50},~\ref{sfig:fix_Ntea_multidom} in the supplemental material).
    \item Table~\ref{stab:raw_ft_fs} shows fine-tuning few-shot results for Figure~\ref{sfig:ft_fs} in the supplemental material.
    \item Table~\ref{stab:raw_ft_ls} shows fine-tuning full-shot results for Figure~\ref{fig:ft_all} in the main paper (Figures~\ref{sfig:ft_ls},~\ref{sfig:ft_ls_full} in the supplemental material.)
\end{itemize}
\clearpage

\begin{table}
    \centering
    \resizebox{\columnwidth}{!}{
    \begin{tabular}{lcccccccccc}\toprule
 & Cars196 & Resisc45 & iFood & \begin{tabular}[c]{@{}c@{}}Food101\\(all)\end{tabular} & \begin{tabular}[c]{@{}c@{}}iFashion\\(sleeve)\end{tabular} & DTD & Flowers & Caltech256 & Birds & Aircrafts \\ \midrule
ImageNet-pretrained model & 31.94 & 70.66 & 28.94 & 36.95 & 89.06 & 60.98 & 83.84 & 80.28 & 54.73 & 30.71 \\
ImageNet-pretrained, soft-label & 31.47 & 69.99 & 29.21 & 37.29 & 89.18 & 60.44 & 82.86 & 81.35 & 57.19 & 29.77 \\ \midrule
Cars196 (t-split) repr. consolid. & 59.21 & 70.00 & 28.42 & 35.28 & 89.00 & 58.99 & 83.01 & 80.26 & 54.02 & 29.85 \\
Cars196 (t-split) trad. distill & 59.39 & 51.56 & 9.75 & 11.54 & 84.27 & 36.07 & 59.91 & 35.51 & 16.90 & 16.82 \\
Resisc45 (t-split) repr. consolid. & 30.80 & 72.64 & 28.77 & 36.37 & 89.03 & 60.08 & 82.75 & 80.77 & 55.47 & 29.17 \\
Resisc45 (t-split) trad. distill & 8.97 & 61.62 & 9.85 & 12.11 & 84.97 & 37.80 & 56.06 & 34.14 & 11.23 & 11.57 \\
iFood (t-split) repr. consolid. & 29.44 & 69.49 & 38.85 & 47.19 & 89.04 & 57.16 & 82.66 & 79.17 & 51.07 & 27.31 \\
iFood (t-split) trad. distill & 12.85 & 53.57 & 35.26 & 44.05 & 85.39 & 36.62 & 70.99 & 39.94 & 17.95 & 14.90 \\
iFashion category task repr. consolid. & 31.89 & 70.02 & 29.23 & 37.45 & 89.15 & 60.76 & 83.28 & 81.29 & 57.10 & 30.24 \\
iFashion category task trad. distill & 6.41 & 37.97 & 6.26 & 8.09 & 90.70 & 23.45 & 44.01 & 30.32 & 9.28 & 9.27 \\
DTD (t-split) repr. consolid. & 31.33 & 69.78 & 29.17 & 37.38 & 89.08 & 61.43 & 82.80 & 81.28 & 56.09 & 29.59 \\
DTD (t-split) trad. distill & 11.29 & 53.12 & 14.11 & 19.00 & 86.39 & 46.80 & 62.48 & 46.38 & 17.14 & 14.75 \\
Flowers (t-split) repr. consolid. & 31.81 & 69.94 & 29.46 & 37.01 & 89.13 & 60.06 & 84.79 & 81.18 & 57.07 & 30.41 \\
Flowers (t-split) trad. distill & 16.72 & 58.08 & 16.68 & 20.67 & 87.03 & 44.70 & 77.35 & 54.49 & 31.32 & 20.26 \\
Caltech256 (t-split) repr. consolid. & 31.71 & 69.36 & 29.18 & 36.97 & 89.08 & 61.07 & 83.14 & 81.28 & 56.63 & 29.98 \\
Caltech256 (t-split) trad. distill & 21.6 & 59.67 & 19.44 & 25.10 & 87.35 & 51.59 & 75.52 & 68.98 & 36.09 & 23.32 \\
CUB (t-split) repr. consolid. & 31.50 & 70.35 & 29.16 & 36.83 & 89.20 & 60.32 & 83.30 & 80.89 & 66.37 & 29.65 \\
CUB (t-split) trad. distill & 17.15 & 57.95 & 17.06 & 20.72 & 87.03 & 46.37 & 72.16 & 53.26 & 62.41 & 19.00 \\
Aircrafts (t-split) repr. consolid. & 30.71 & 69.58 & 28.25 & 35.54 & 89.06 & 58.67 & 81.36 & 80.54 & 53.95 & 55.26 \\
Aircrafts (t-split) trad. distill & 10.56 & 44.30 & 7.81 & 9.09 & 84.17 & 29.16 & 48.32 & 29.18 & 10.87 & 58.51 \\
iNaturalist (t-split) repr. consolid. & 30.08 & 69.67 & 27.35 & 34.52 & 88.70 & 57.28 & 84.77 & 78.01 & 67.58 & 27.94 \\
iNaturalist (t-split) trad. distill & 13.51 & 57.25 & 14.66 & 18.68 & 84.46 & 39.49 & 79.74 & 41.59 & 57.49 & 16.98 \\  \midrule
Cars196 10\% x 5 (Res18) repr. consolid. & 42.03 & 69.95 & 29.44 & 36.97 & 89.05 & 60.15 & 83.23 & 81.48 & 58.18 & 31.10 \\
Cars196 10\% x 5 (Res18) trad. distill & 45.37 & 60.74 & 17.53 & 20.80 & 87.11 & 48.16 & 73.66 & 58.56 & 39.02 & 23.85 \\
Cars196 10\% (Res18) repr. consolid. & 36.97 & 69.64 & 28.62 & 35.95 & 88.92 & 59.88 & 82.79 & 80.37 & 56.80 & 30.87 \\
Cars196 10\% classes (Res18) teacher & 34.28 & 56.03 & 16.83 & 20.78 & 87.14 & 45.59 & 70.31 & 57.13 & 39.40 & 22.33 \\
Cars196 10\% classes (Res18) teacher (best of 5) & 35.39 & 56.47 & 16.98 & 20.50 & 87.59 & 45.77 & 70.58 & 60.41 & 41.87 & 23.40 \\
iFood 10\% x 5 (Res18) repr. consolid. & 31.14 & 70.69 & 35.82 & 42.97 & 88.98 & 58.98 & 83.51 & 80.70 & 53.91 & 29.80 \\
iFood 10\% x 5 (Res18) trad. distill & 17.60 & 60.26 & 34.10 & 39.71 & 86.58 & 45.93 & 76.18 & 51.55 & 27.91 & 20.27 \\
iFood 10\% (Res18) repr. consolid. & 30.45 & 69.54 & 31.25 & 38.47 & 88.99 & 58.59 & 81.97 & 79.37 & 54.33 & 30.25 \\
iFood 10\% classes (Res18) teacher & 13.77 & 52.88 & 22.29 & 25.46 & 85.84 & 36.14 & 65.34 & 45.19 & 21.58 & 16.43 \\
iFood 10\% classes (Res18) teacher (best of 5) & 14.27 & 52.35 & 23.11 & 26.40 & 86.45 & 38.34 & 65.88 & 44.94 & 22.58 & 16.11 \\  \midrule
Cars196 10\% x 5 repr. consolid. & 45.99 & 70.10 & 29.23 & 37.02 & 89.16 & 60.40 & 83.82 & 81.21 & 57.40 & 31.42 \\
Cars196 10\% x 5 trad. distill & 50.33 & 62.63 & 18.52 & 22.32 & 86.77 & 49.31 & 75.66 & 59.74 & 37.27 & 25.48 \\
Cars196 10\% repr. consolid. & 39.23 & 69.62 & 28.42 & 36.01 & 89.03 & 59.84 & 82.76 & 80.34 & 56.43 & 31.20 \\
Cars196 10\% classes teacher & 41.76 & 60.72 & 19.63 & 24.53 & 87.85 & 50.03 & 78.77 & 61.27 & 39.22 & 23.27 \\
Cars196 10\% classes teacher (best of 5) & 42.92 & 64.48 & 20.87 & 25.95 & 87.58 & 51.79 & 78.69 & 64.35 & 43.20 & 25.01 \\
iFood 10\% x 5 repr. consolid. & 30.74 & 70.64 & 37.34 & 44.77 & 88.91 & 59.17 & 84.58 & 80.20 & 52.47 & 29.83 \\
iFood 10\% x 5 trad. distill & 17.70 & 61.33 & 35.88 & 42.71 & 86.18 & 44.63 & 77.50 & 50.98 & 25.36 & 20.49 \\
iFood 10\% repr. consolid. & 30.05 & 68.62 & 32.06 & 39.37 & 88.83 & 58.67 & 82.09 & 79.43 & 53.83 & 29.68 \\
iFood 10\% classes teacher & 14.07 & 52.84 & 25.04 & 28.84 & 86.70 & 41.14 & 69.59 & 45.79 & 19.57 & 16.57 \\
iFood 10\% classes teacher (best of 5) & 13.85 & 54.31 & 25.16 & 29.22 & 86.48 & 39.39 & 69.26 & 44.21 & 20.54 & 16.08 \\  \midrule
Cars196 + Resisc45 + iFood (t-split) repr. consolid. & 47.85 & 72.39 & 34.57 & 41.92 & 89.09 & 58.63 & 83.64 & 79.90 & 51.79 & 29.18 \\
Cars196 + Resisc45 + iFood (t-split) trad. distill & 53.27 & 66.27 & 34.44 & 41.84 & 85.32 & 43.74 & 75.84 & 48.16 & 23.16 & 19.99 \\
\begin{tabular}[c]{@{}r@{}}Cars196 + Resisc45 + iFood (t-split) repr. consolid. \\ (Places+iNat+logo2k+ImNet proxy 10 epochs)\end{tabular} & 45.71 & 73.00 & 32.87 & 39.80 & 89.00 & 59.09 &83.89 & 79.83 & 52.40 & 29.46 \\
\begin{tabular}[c]{@{}r@{}}Cars196 + Resisc45 + iFood (t-split) trad. distill \\ (Places+iNat+logo2k+ImNet proxy 10 epochs)\end{tabular} & 53.26 & 66.34 & 32.80 & 39.10 & 85.43 & 43.81 & 75.92 & 47.75 & 23.15 & 20.15 \\  \midrule
Cars196 (t-split) repr. consolid. old:new = 1:3 & 61.21 & 67.83 & 26.70 & 33.35 & 88.92 &  &  &  &  &  \\
Cars196 (t-split) repr. consolid. old:new = 1:1 & 59.21 & 70.00 & 28.42 & 35.28 & 89.00 &  &  &  &  &  \\
Cars196 (t-split) repr. consolid. old:new = 3:1 & 52.80 & 70.82 & 29.18 & 36.53 & 89.13 &  &  &  &  &  \\
Cars196 (t-split) repr. consolid. (ImageNet proxy) & 59.21 & 70.00 & 28.42 & 35.28 & 89.00 &  &  &  &  &  \\
Cars196 (t-split) repr. consolid. (Places365 proxy) & 61.18 & 69.59 & 26.69 & 33.54 & 88.68 &  &  &  &  &  \\
Cars196 (t-split) trad. distill (Places365 proxy) & 59.98 & 51.11 & 9.54 & 11.10 & 84.38 &  &  &  &  &  \\
\begin{tabular}[c]{@{}r@{}}Cars196 (t-split) repr. consolid. \\ (Places365 proxy w/ label)\end{tabular} & 53.16 & 69.32 & 23.17 & 28.54 & 88.23 &  &  &  &  &  \\  \bottomrule
\end{tabular}
}
    \caption{Fixed representation, few-shot results raw numbers for Figures~\ref{fig:fix_1tea},~\ref{fig:fix_Ntea},~\ref{fig:side} in the main paper (Figures~\ref{sfig:fix_1tea},~\ref{sfig:fix_Ntea_r18},~\ref{sfig:fix_Ntea_r50},~\ref{sfig:fix_Ntea_multidom} in the supplemental material)}
    \label{stab:raw_fix}
\end{table}

\begin{table}
    \centering
    \resizebox{\columnwidth}{!}{
    \begin{tabular}{lccccc} \toprule
  & Cars196 & Resisc45 & iFood & \begin{tabular}[c]{@{}c@{}}Food101\\(all)\end{tabular} & \begin{tabular}[c]{@{}c@{}}iFashion\\(sleeve)\end{tabular} \\ \midrule
ImageNet-pretrained model & 37.18 & 68.91 & 28.25 & 34.72 & 89.68 \\
ImageNet-pretrained, soft-label & 39.84 & 69.88 & 28.72 & 35.31 & 89.92 \\ \midrule
Cars196 (t-split) repr. consolid. & 55.80 & 69.91 & 28.41 & 34.00 & 89.91 \\
Cars196 (t-split) trad. distill & 60.19 & 61.30 & 18.98 & 19.79 & 87.90 \\
Resisc45 (t-split) repr. consolid. & 38.30 & 71.71 & 28.16 & 34.33 & 89.83 \\
Resisc45 (t-split) trad. distill & 21.12 & 60.93 & 17.57 & 19.46 & 88.05 \\
iFood (t-split) repr. consolid. & 35.94 & 68.39 & 35.27 & 42.14 & 89.54 \\
iFood (t-split) trad. distill & 27.89 & 61.58 & 31.23 & 39.33 & 87.91 \\
iFashion category task repr. consolid. & 38.57 & 68.45 & 28.80 & 35.88 & 89.29 \\
iFashion category task trad. distill & 16.71 & 54.53 & 11.49 & 14.93 & 89.80 \\ \midrule
Cars196 + Resisc45 + iFood (t-split) repr. consolid. & 45.31 & 70.84 & 30.90 & 37.62 & 89.87 \\
Cars196 + Resisc45 + iFood (t-split) trad. distill & 45.29 & 67.05 & 30.14 & 35.44 & 87.97 \\ \bottomrule
\end{tabular}
}
    \caption{Fine-tuning, few-shot results raw numbers for Figure~\ref{sfig:ft_fs} in the supplemental material}
    \label{stab:raw_ft_fs}
\end{table}

\begin{table}
    \centering
    \resizebox{\columnwidth}{!}{
    \begin{tabular}{lccccccc} \toprule
 & Cars196 & Resisc45 & iFood & \begin{tabular}[c]{@{}c@{}}Food101\\(all)\end{tabular} & Aircrafts & \begin{tabular}[c]{@{}c@{}}Flowers\\(all)\end{tabular} & \begin{tabular}[c]{@{}c@{}}Birds\\(all)\end{tabular} \\  \midrule
ImageNet-pretrained model & 90.90 & 96.93 & 78.55 & 88.15 & 87.57 & 92.32 & 80.82 \\
ImageNet-pretrained, soft-label & 90.85 & 97.20 & 78.92 & 88.13 & 87.87 & 92.54 & 81.76 \\ \midrule
Cars196 (t-split) repr. consolid. & 91.84 & 97.08 & 78.12 & 87.91 & 87.81 & 92.45 & 81.46 \\
Cars196 (t-split) trad. distill & 91.35 & 96.43 & 75.76 & 86.10 & 86.91 & 85.35 & 75.39 \\
Resisc45 (t-split) repr. consolid. & 91.03 & 97.14 & 78.57 & 88.02 & 87.75 & 92.08 & 81.55 \\
Resisc45 (t-split) trad. distill & 87.41 & 96.71 & 75.90 & 86.09 & 85.42 & 85.25 & 71.57 \\
iFood (t-split) repr. consolid. & 90.75 & 96.86 & 78.34 & 88.00 & 87.75 & 92.26 & 80.20 \\
iFood (t-split) trad. distill & 89.44 & 96.34 & 77.24 & 87.33 & 85.65 & 88.42 & 76.30 \\
iFood (t-split) teacher & 89.09 & 96.21 & 77.76 & 87.92 & 84.46 & 87.97 & 75.92 \\
Aircrafts (t-split) repr. consolid. & 90.93 & 97.08 & 78.20 & 88.01 & 88.94 & 91.97 & 81.15 \\
Aircrafts (t-split) trad. distill & 87.31 & 95.87 & 75.06 & 85.29 & 89.54 & 81.79 & 72.30 \\
iNaturalist (t-split) repr. consolid. & 91.08 & 96.74 & 78.16 & 87.92 & 87.81 & 93.62 & 82.36 \\
iNaturalist (t-split) trad. distill & 88.62 & 96.02 & 76.64 & 86.60 & 86.61 & 90.89 & 79.89 \\
iNaturalist (t-split) teacher & 89.37 & 96.37 & 76.89 & 87.09 & 85.59 & 91.22 & 79.62 \\ \midrule
Cars196 + Resisc45 + iFood (t-split) repr. consolid. & 91.05 & 96.74 & 78.42 & 88.15 & 88.16 & 92.76 & 81.43 \\
Cars196 + Resisc45 + iFood (t-split) trad. distill & 91.05 & 96.71 & 77.47 & 87.38 & 87.27 & 89.71 & 77.29 \\ \bottomrule
\end{tabular}
}
    \caption{Fine-tuning, full-shot results raw numbers for Figure~\ref{fig:ft_all} in the main paper (Figures~\ref{sfig:ft_ls},~\ref{sfig:ft_ls_full} in the supplemental material.}
    \label{stab:raw_ft_ls}
\end{table}

\end{document}